\documentclass{article}

    \PassOptionsToPackage{numbers, compress}{natbib}

\usepackage[final]{neurips_2021}

\usepackage[utf8]{inputenc} 
\usepackage[T1]{fontenc}    
\usepackage{hyperref}       
\usepackage{url}            
\usepackage{booktabs}       
\usepackage{amsfonts}       
\usepackage{nicefrac}       
\usepackage{microtype}      
\usepackage{xcolor}         

\usepackage{amsmath,amsfonts,bm, bbm}

\def\eqref#1{equation~\ref{#1}}

\def\1{\bm{1}}

\DeclareMathAlphabet{\mathsfit}{\encodingdefault}{\sfdefault}{m}{sl}
\SetMathAlphabet{\mathsfit}{bold}{\encodingdefault}{\sfdefault}{bx}{n}

\newcommand{\R}{\mathbb{R}}
\newcommand{\one}{\mathbbm{1}}

\usepackage{graphicx}
\usepackage{caption}
\usepackage{subcaption}
\usepackage{makecell}
\usepackage{multirow}

\usepackage{widetable}

\usepackage[ruled,vlined]{algorithm2e}

\newcolumntype{C}[1]{>{\centering\arraybackslash}m{#1}}

\newcommand*{\affaddr}[1]{#1} 
\newcommand*{\affmark}[1][*]{\textsuperscript{#1}}

\title{Predicting Deep Neural Network Generalization\\with Perturbation Response Curves}

\author{
Yair Schiff\affmark[1], Brian Quanz\affmark[2], Payel Das\affmark[2], Pin-Yu Chen\affmark[2] \\
\affaddr{\affmark[1]IBM Watson}, \affaddr{\affmark[2]IBM Research}\\
\texttt{\{yair.schiff@,blquanz@us.,daspa@us.,pin-yu.chen@\}ibm.com}
}

\begin{document}

\maketitle

\begin{abstract}
The field of Deep Learning is rich with empirical evidence of human-like performance on a variety of prediction tasks.
However, despite these successes, the recent Predicting Generalization in Deep Learning (PGDL) NeurIPS 2020 competition \cite{jiang2020neurips} suggests that there is a need for more robust and efficient measures of network generalization.
In this work, we propose a new framework for evaluating the generalization capabilities of trained networks.
We use perturbation response (PR) curves that capture the accuracy change of a given network as a function of varying levels of training sample perturbation.
From these PR curves, we derive novel statistics that capture generalization capability.
Specifically, we introduce two new measures for accurately predicting generalization gaps: the Gi-score and Pal-score, which are inspired by the Gini coefficient and Palma ratio (measures of income inequality), that accurately predict generalization gaps.
Using our framework applied to \textit{intra} and \textit{inter}-class sample mixup, we attain better predictive scores than the current state-of-the-art measures on a majority of tasks in the PGDL competition.
In addition, we show that our framework and the proposed statistics can be used to capture to what extent a trained network is invariant to a given parametric input transformation, such as rotation or translation.
Therefore, these generalization gap prediction statistics also provide a useful means for selecting optimal network architectures and hyperparameters that are invariant to a certain perturbation.
\end{abstract}

\section{Introduction}\label{sec:intro}
Neural networks have produced state-of-the-art and human-like performance across a variety of tasks.
This rapid progress has led to wider-spread adoption and deployment.
Given their prevalence and increasing applications, it is important to estimate how well a trained net will generalize.
Additionally, specific tasks often require models to be invariant to certain transformations or perturbations of the data.
This can be achieved either through data augmentation that changes the underlying statistics of the training sets or through inductive architectural biases, such as translation invariance that is inherent in convolutional neural networks.
It is important as well to understand how and when a network has been able to learn task-dependent invariances. 

Various attempts at bounding and predicting neural network generalization are well summarized and analyzed in the recent survey \cite{jiang2019fantastic}.
While both theoretical and empirical progress has been made, there remains a gap in the literature for an efficient and intuitive measure that can predict generalization given a trained network and its corresponding data \textit{post hoc}.
Aiming to fill this gap, the recent Predicting Generalization in Deep Learning (PGDL) NeurIPS 2020 competition \cite{jiang2020neurips} encouraged participants to provide \textit{complexity} measures that would take into account network weights and training data to predict generalization gaps, i.e., the difference between training and test set accuracy.

In this work, we propose a new framework that presents progress towards this goal.
Our methodology consists of first building an estimate of how the accuracy of a network changes as a function of varying levels of perturbation present in training samples.
To do so, we evaluate a trained network's accuracy on a subset of the training dataset that has been perturbed to some degree.
Using multiple observations of accuracy vs. perturbation magnitude, we develop a perturbation response (PR) curve for each model.
From the PR curves, we derive two new measures called the Gi-score and the Pal-score, which compare a given network's PR curve to that of an idealized network that is unaffected by all perturbation magnitudes.
When applying our framework to \textit{inter} and \textit{intra} class Mixup \cite{zhang2017mixup} perturbations, we are able to achieve better generalization prediction scores on a majority of the tasks than the current state-of-the-art proposal from the PGDL competition.
Because our framework can be applied to any parametric perturbation, we also demonstrate how it can be used to predict the degree to which a network has learned to be invariant to a given perturbation.

\section{Related work}\label{sec:related_work}
The PGDL competition resulted in several proposals of complexity measures that aim to predict neural network generalization gaps.
While several submissions build off the work of \cite{jiang2018predicting} and rely on margin-based measures, we will focus on those submissions that measure perturbation response, specifically to Mixup, since this is most relevant to our work.
Mixup, first introduced in \cite{zhang2017mixup}, is a novel training paradigm in which training occurs not just on the given training data, but also on linearly interpolated points.
Manifold Mixup training extends this idea to interpolation of intermediate network representations \cite{verma2019manifold}.

Perhaps most closely related to our work is that of the winning submission, \cite{natekar2020representation}.
While \cite{natekar2020representation} presents several proposed complexity measures, the authors explore accuracy on Mixup and Manifold Mixup training sets as potential predictors of generalization gap, and performance on mixed up data is one of the inputs into their winning submission.
While this closely resembles our work, in that the authors are using performance on a perturbed dataset, namely a mixed up one, the key difference is that \cite{natekar2020representation} only investigates a network's response to a single magnitude of interpolation, 0.5.
Additionally, we investigate \textit{between}-class interpolation as well, while \cite{natekar2020representation} only interpolates between data points or intermediate representations within a class, not between classes.
Our proposed Gi-score and Pal-score therefore provide a much more robust sense for how invariant a network is to this mixing up perturbation and can easily be applied to other perturbations as well.

In the vein of exploring various transformations / perturbations, the second place submission \cite{kashyap2021robustness} performs various augmentations, such as color saturation, applying Sobel filters, cropping and resizing, and others, to create a composite penalty score based on how a network performs on these perturbed data points.
Our work, in addition to achieving better generalizatiton gap prediction scores, can be thought of as an extension of \cite{kashyap2021robustness}, because as above, rather than looking at a single perturbation level, the Gi-score and Pal-score provide a summary of how a model reacts to a spectrum of parameterized perturbations.

In this work, we also extend our proposed Gi and Pal-scores to predict generalization performance on different data transformations.
\cite{azulay2018deep} shows that even for architectures that have inductive biases that should render a network invariant to certain perturbations, in practice these networks are not actually invariant due to how they are trained.
This is true when data augmentation is not carefully employed \cite{azulay2018deep}, which highlights the need for our predicting invariance line of work.
Unlike works, such as \cite{goodfellow2009measuring}, that measure how individual layer and neuron outputs respond to input perturbations, e.g., rotation, translation, and color-jittering, we measure how a network's overall accuracy responds to the these perturbations and compare that to an idealized network that is fully invariant.
Consistent prediction on invariant data transformations is a desired property for neural networks \cite{bronstein2021geometric} and it has been widely used as a data augmentation or regularization tool during training for improving generalization \cite{benton2020learning}.

Our scores are inspired by the Gini coefficient and Palma ratio - most commonly used in economics as measures of income inequality \cite{lorenz1905methods,cobham2013all,dagum2001encyclopedia,lambert1993inequality}.
In economics, the Gini coefficient measures income inequality by ordering a population by income and plotting the percentage of the total national income on the vertical axis vs. the percentage of the population total on the horizontal axis.
For an \textit{idealized} economy, wealth distribution lies on a $45^{\circ}$ line from the origin, which means each percent of the population holds the same percentage of wealth.
Plotting this distribution for an actual economy allows for the calculation of the Gini coefficient by taking a ratio of the area between the idealized and actual economy wealth distribution curves and the total area below the idealized economy's curve.
The Palma ratio is also calculated from this plot by taking a ratio of some top $x\%$ of the population's wealth (area below the actual economy's curve) divided by that of some bottom $y\%.$

In addition to income inequality, the Gini coefficient has been used in a wide variety of applications in different domains, e.g, measuring biodiversity in ecology \cite{wittebolle2009initial}, quality of life in health \cite{asada2005assessment}, and protein selectivity in chemistry \cite{graczyk2007gini}.
It is also used in machine learning for measuring classification performance, as twice the area between the ROC curve (curve of false positive rate vs. true positive rate for varying classifier threshold) and the diagonal \cite{hand2001simple,bekkar2013evaluation} and has also been used as a criteria for feature selection \cite{sanasam10a}.
Our approach is the first to use this metric together with our PR curve framework for measuring generalization, as well as to use the Pal-score in a machine learning setting.


\section{Methodology}\label{sec:methodology}
\subsection{Notation}\label{subsec:notation}
We begin by defining a network for a classification task as $f: \R^d \rightarrow \Delta_k$; that is, a mapping of real input signals $x$ of dimension $d$ to discrete distributions, with $\Delta_k$ being the space of all $k$-simplices.
We also define the intermediate layer mappings of a network as $f^{(\ell)}: \R^{d_{\ell-1}} \rightarrow \R^{d_{\ell}}$, where $\ell$ refers to a layer's depth with dimension $d_{\ell}$.
The output of each layer is defined as $x^{(\ell)} = f^{(\ell)}(x^{(\ell-1)})$, with inputs defined as $x^{(0)}$.
Additionally, let $f_{\ell}: \R^{d_\ell} \rightarrow \Delta_k$ be the function that maps intermediate representations $x^{(\ell)}$ to the final output of probability distributions over classes. 
For a dataset $\mathcal{D}$, consisting of pairs of inputs $x \in \R^d$ and labels $y \in [k]$, a network's accuracy is defined as $\mathcal{A} = \sum_{x, y \in \mathcal{D}}\one(\max_{i \in [k]}f(x)[i] = y) \ / \ |\mathcal{D}|,$
i.e. the fraction of samples where the predicted class matches the ground truth label, where $\one(\cdot)$ is an indicator function and $f(x)[i]$ refers to the probability weight of the $i^{\mathrm{th}}$ class.

We define perturbations of the network's representations as $\mathcal{T}_{\alpha}: \R^{d_{\ell}} \rightarrow \R^{d_{\ell}}$, where $\alpha$ controls the magnitude of the perturbation.
For example, changing the intensity of an image by $\alpha$ percent can be represented as $\mathcal{T}_\alpha(x^{(0)}) = \alpha x^{(0)}.$
To measure a network's response to a perturbation $\mathcal{T}_\alpha$ applied at the $\ell^{\mathrm{th}}$ layer output, we calculate the accuracy of the network for a sample of the training data on which the perturbation has been applied:
\begin{equation}
\label{eq:perturb_acc}
\mathcal{A}_\alpha^{(\ell)} = \sum_{x, y \sim \mathcal{D}_{sample}}\one(\max_{i \in [k]}f_{\ell}(\mathcal{T}_\alpha(x^{(\ell)}))[i] = y) \ / \ |\mathcal{D}_{sample}|.
\end{equation}
The greater the gap $\mathcal{A} - \mathcal{A}_\alpha^{(\ell)}$, the less the network is resilient or invariant to the perturbation $\mathcal{T}_\alpha$ when applied to the $\ell^{\mathrm{th}}$ layer.  Perturbations at deeper network layers can be viewed as perturbations in an implicit feature space learned by the network.

\subsection{Calculating Perturbation Response curves}\label{subsec:pr_curves}
\begin{figure}[h]
    \centering
    \includegraphics[width=0.45\linewidth]{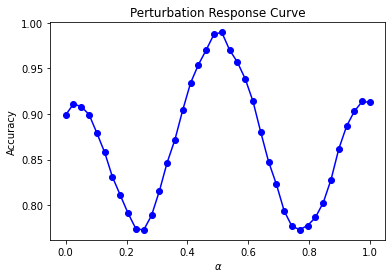}
    \includegraphics[width=0.45\linewidth]{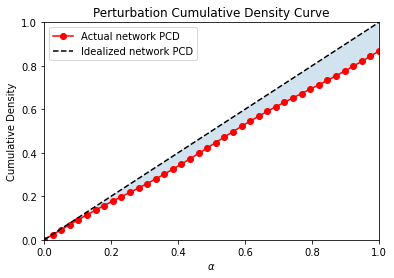}
    \caption{(Left) Sample perturbation response (PR) curve displaying a networks accuracy on 10\% of the training data at varying magnitudes of a perturbation.
    (Right) Sample perturbation cumulative density (PCD) curve comparing the actual network's cumulative density (red line) to that of an idealized network (dotted black line) with shaded area between the curves, which is used in Gi-score calculations.
    Pal-score comes from the area below the red curve.
    These sample plots come from a Resnet \cite{he2016deep} model trained on SVHN \cite{netzer2011reading} evaluated with rotation as the parameterized perturbation.
    $\alpha$ varies from $-90^{\circ}$ to $90^{\circ}$ of rotation and is normalized in the plots to range from 0 to 1.}
    \label{fig:pr_pcd_curve}
\end{figure}

To measure a network's robustness to a perturbation, one could simply choose a fixed $\alpha$ and measure the network's response.
However, a more complete picture is provided by sampling the network's response to various magnitudes of $\alpha$.
In Figure \ref{fig:pr_pcd_curve}, we show this in practice.
For example, letting $\mathcal{T}_\alpha$ refer to image rotation, we vary $\alpha$ from a minimum $\alpha_{\min}$ degree of rotation, to a maximum $\alpha_{\max}$ degree.
For each $\alpha$, we calculate accuracy  $\mathcal{A}_\alpha^{(\ell)}$ to measure the network's response to the perturbation of magnitude $\alpha$ applied at depth $\ell$.
Plotting $\mathcal{A}_\alpha^{(\ell)}$ on the vertical axis and $\alpha$ on the horizontal axis gives us the PR curves.
In Figure \ref{fig:pr_pcd_curve}, we display rotation applied to training images ($\ell = 0$) from the SVHN dataset.
This methodology is summarized in Algorithm \ref{alg:pr_curve} in Appendix \ref{app:alg_prc}.

\subsection{Calculating the Gi-score and Pal-score}\label{subsec:gi_score}
To extract a single statistic from the PR curves in Figure \ref{fig:pr_pcd_curve}, we draw inspiration from the Gini coefficient and Palma ratio.
Namely, we compare a network's response to varying magnitudes of perturbations with an \textit{idealized} network: one whose accuracy is unaffected by the perturbations.
The idealized network therefore has a PR curve that starts and remains at accuracy 1.0.

This comparison is achieved by creating a new graph that plots the cumulative density integral under the PR curves against the magnitudes $\alpha_i \in [\alpha_{\min}, \alpha_{\max}]$:
$\int_{0}^{\alpha_i} \mathcal{A}_\alpha d\alpha$.
This produces what we call \textit{perturbation cumulative density} (PCD) curves seen in Figure \ref{fig:pr_pcd_curve}.
For the idealized network whose PR is identically equal to 1 for all $\alpha$, this PCD curve is just the $45^{\circ}$ line passing through the origin.
Finally, the \textbf{Gi-score} (named for the Gini coefficient it draws inspiration from) is calculated by taking a ratio of the area between the idealized network's PCD curve and that of the actual network and the total area below the idealized network PCD.
We summarize this in Algorithm \ref{alg:gi_score} in Appendix \ref{app:alg_gini}.

The \textbf{Pal-score} (named for the Palma ratio it draws inspiration from) calculates the area under the PCD curve and takes a ratio of the area for top 60\% of perturbation magnitudes divided by the area for the bottom 10\%.
This allows us to focus on variations on the upper and lower ends of the perturbation magnitude spectrum, ignoring the middle perturbations that might not vary as widely across networks.
We give the pseudocode for the Pal-score in Algorithm \ref{alg:pal_score} in Appendix \ref{app:alg_pal}.

\subsection{Motivation}\label{subsec:motivation}
Before presenting the empirical effectiveness of our work, we provide detailed intuition underlying our framework.
In Economics, the Gini coefficient is specifically designed to compare how the wealth distribution of a given economy compares to that of an idealized economy, where each percentage of the population has an equal share of the nation's income.
In that way, the Gini coefficient does not characterize a nation's inequality by simply looking at the percentage of income that an individual percentile of the population holds, rather, the Gini coefficient extracts aggregate information from the entire Lorenz curve which plots the wealth distribution across all percentiles of the population.
In our work, this was the main inspiration for creating the Gi-score.
We can measure a neural network's response to a certain magnitude of a transformation, but \textit{a priori}, we do not know on which magnitude to focus.
We hypothesize that looking at a single value of perturbation magnitude, which is similar to the Mixup approach in \cite{natekar2020representation}, may not be ideal and/or informative, and that this can be viewed as a specific case of the more general framework we propose here: evaluating a whole curve of perturbation magnitudes.
We therefore propose to examine the spectrum of perturbation-responses and extract a more holistic view of how resilient a network is to a given transformation. 
A key reasoning behind this is that even if two networks have the same accuracy after the maximum amount of perturbation (e.g., 0.5 for the case of Mixup) that could be a final base level of accuracy deterioration -- that multiple networks may even share -- but the amount of deterioration before that point could still be different for different networks, and this may also be predictive of generalization behavior (Figure \ref{fig:pr_curve_compare}).

Justification for using a Pal-score is similar to the reasoning provided above.
However, there may be uses cases or domains where there are certain magnitudes of perturbation that are consistent across models, and are therefore uninformative, leading one to focus on certain regions of the PR curves, as with the Palma ratio.

  
        

\section{Experiments}\label{sec:exps}
\subsection{Generalization predictions}\label{subsec:gen_pred}
To evaluate the extent to which the Gi and Pal-scores accurately predict network generalization, we calculate these statistics on the corpus of trained models provided in the PGDL competition \cite{jiang2020neurips}.
We use the trained networks and their configurations, training data, and ``starting kit'' code from the competition; all open-sourced and provided under Apache 2.0 license\footnote{https://github.com/google-research/google-research/tree/master/pgdl}.
The code includes utilities for loading models and model details and running scoring.
To this base repository, we added our methods for performing different perturbations at different layers, computing PR curves, and computing our proposed Gi and Pal-scores.
Note, we focus on the image modality as this is the type of input data available in the PGDL competition and has been the primary data modality used in prior related work on predicting generalization in other setups as well \cite{jiang2019fantastic,jiang2018predicting,forouzesh2021generalization}. 
Future work will evaluate the application of our framework to other modalities.

\subsubsection{Generalization predictions: Experimental setup}\label{subsubsec:gen_pred_exp}
The networks from this competition contain the following architectures: VGG \cite{simonyan2014very}, Network-in-Network (NiN) \cite{lin2013network}, and Fully Convolutional \cite{lecun1995convolutional} (Conv) architectures.
The datasets are comprised of CIFAR-10 \cite{krizhevsky2009learning}, SVHN \cite{netzer2011reading}, CINIC-10 \cite{darlow2018cinic}, Oxford Flowers \cite{nilsback2008automated}, Oxford Pets \cite{parkhi2012cats}, and Fashion MNIST \cite{xiao2017fashion}.
Note, to our knowledge, these datasets are not known to contain personally identifiable information or offensive content.
Although CIFAR-10 and CINIC-100 use images from  the problematic ImageNet and Tiny Images \cite{prabhu2020large}, they contain manually selected subsets.
The list of dataset-model combinations, or tasks, available in the trained model corpus can be seen in the first two rows of Table \ref{tab:gen_pred_results}.
Across the 8 tasks, there are a total of 550 networks.
Each network was trained so that it attains nearly perfect accuracy on the training dataset.

As proposed in \cite{jiang2020neurips}, the goal is to find a \textit{complexity} measure of networks that is causally informative (predictive) of generalization gaps.
To measure this predictive quality, \cite{jiang2020neurips} propose a Conditional Mutual Information (CMI) score, 
which measures how informative the complexity measure is about the network's generalization given the network's hyperparameters (i.e., the information contributed by the measure in addition to the network hyperparameters).
For full implementation details of this score, please refer to \cite{jiang2020neurips} and see Appendix \ref{app:cmi}, but roughly, higher values of CMI represent greater capability of a complexity score in predicting generalization gaps.

In our experiments, we let $\mathcal{T}_\alpha$ be defined as an interpolation between two points of either different or the same class:
$\mathcal{T}_\alpha(x) = (1-\alpha)x + \alpha x'$,
For \textit{inter}-class interpolation, where $x'$ is a (random) input from a different class than $x$, we range $\alpha \in [0, 0.5)$.
For the \textit{intra}-class setup, where $x$ and $x'$ are drawn from the same class, we include the upper bound of the magnitude: $\alpha \in [0, 0.5].$
While we explored other varieties of perturbation, such as adding Gaussian noise, we found that this interpolation perturbation was most predictive of generalization gaps for the networks and datasets we tested.
Both interpolation perturbations that we test (intra and inter-class) are justifiable for predicting generalization gap.
We hypothesize that invariance to interpolation \textit{within} a class should indicate that a network produces similar representations and ultimately the same class maximum prediction for inputs and latent representations that are within the same class regions.
Invariance to interpolation \textit{between} classes up to 50\% should indicate that the network has well separated clusters for representations of different classes and is robust to perturbations moving points away from heavily represented class regions in the data.

\subsubsection{Generalization predictions: Results}\label{subsubsec:gen_pred_results}
In Table \ref{tab:gen_pred_results}, we present average CMI scores for all models within a task for our Gi and Pal-scores compared to that of the winning team \cite{natekar2020representation} from the PGDL competition.
We also compare our statistic to comparable ones presented in \cite{natekar2020representation} that rely on Mixup and Manifold Mixup accuracy\footnote{Scores from \cite{natekar2020representation}
if reported, otherwise we use the author-provided code: https://github.com/parthnatekar/pgdl}.
The winning submission described in \cite{natekar2020representation} uses a combination of a score based on the accuracy of mixed up input data and a clustering quality index of class representations, known as the Davies-Bouldin Index (DBI) \cite{davies1979cluster}.
Using the notation introduced in Section \ref{sec:methodology}, the measures from \cite{natekar2020representation} present in Table \ref{tab:gen_pred_results} can be described as follows: Mixup accuracy: $\mathcal{A}_{0.5}^{(0)}$; Manifold Mixup accuracy: $\mathcal{A}_{0.5}^{(1)}$; DBI * Mixup: $DBI * (1-\mathcal{A}_{0.5}^{(0)}).$


\begin{table}[ht!]
\small
\caption{Comparison of Conditional Mutual Information scores for various complexity measures across tasks.
We present single-measure scores in the top part of the table and scores based on combinations of multiple measures in the bottom part.
For each section, the highest score within a task is bolded.
Best scores overall are marked with an asterisk.
In CINIC-10 columns, `bn' stands batch-norm.
}
\label{tab:gen_pred_results}
\begin{center}
\begin{widetabular}{\textwidth}{l | c c | c | c c | c | c | c | c}
\toprule 
\multicolumn{1}{c|}{}
&\multicolumn{2}{c|}{CIFAR-10}
&\multicolumn{1}{c|}{SVHN}
&\multicolumn{2}{c|}{CINIC-10}
&\multicolumn{1}{c|}{\makecell{Oxford \\ Flowers}}
&\multicolumn{1}{c|}{\makecell{Oxford \\Pets}}
&\multicolumn{1}{c|}{\makecell{Fashion \\ MNIST}}
&\makecell[t]{\textit{All} \\ \textit{Avg}}
\\\cline{2-9}
\multicolumn{1}{c|}{}
&\textit{VGG}
&\textit{NiN}
&\textit{NiN}
&\makecell{\textit{Conv}\\\textit{w/bn}}
&\makecell{\textit{Conv}\\}
&\textit{NiN}
&\textit{NiN}
&\textit{VGG}
\\ \midrule
\multicolumn{5}{l}{\textit{Single measures only}}\vspace{3pt}\\
Gi \textit{inter} $\ell$=0    & 3.03         & \bf34.34 & 26.58        & 21.01        & 6.96         & 33.05        & \bf18.46$^*$ & 4.48         & 18.49        \\ 
Gi \textit{inter} $\ell$=1    & \bf7.88  & 22.59        & 12.17        & 12.58        & 8.39         & 7.52         & 4.68         & \bf16.16$^*$ & 11.49        \\ 
Pal \textit{inter} $\ell$=0   & 3.14         & 26.39        & 24.25        & 21.11        & 6.37         & 29.62        & 15.96        & 4.21         & 16.38        \\ 
Pal \textit{inter} $\ell$=1   & 7.31         & 12.75        & 9.79         & 12.09        & 7.71         & 6.37         & 3.46         & 14.13        & 9.20         \\ 
Gi \textit{intra} $\ell$=0    & 0.84         & 30.54        & \bf41.75$^*$ & 22.97        & 11.46        & \bf42.44     & 16.21        & 5.10         & \bf21.41     \\ 
Gi \textit{intra} $\ell$=1    & 0.22         & 17.18        & 10.96        & 9.50         & 12.43        & 6.92         & 3.60         & 5.55         & 8.29         \\ 
Pal \textit{intra} $\ell$=0   & 0.61         & 24.36        & 31.82        & 24.15        & 11.01        & 38.10        & 14.04        & 5.12         & 18.65        \\ 
Pal \textit{intra} $\ell$=1   & 0.44         & 10.34        & 13.48        & 8.68         & 11.09        & 5.88         & 3.02         & 6.25         & 7.40         \\ 
\midrule
Mixup                        & 0.03         & 14.18        & 22.75        & \bf30.30 & \bf19.51 & 35.30        & 9.99         & 7.75         & 17.48        \\ 
Mani. Mix.                  & 2.24         & 2.88         & 12.11        & 4.23         & 4.84         & 0.03         & 0.13         & 0.19         & 3.33         \\ 
\midrule \midrule
\multicolumn{5}{l}{\textit{Combination measures}}\vspace{3pt}\\
PCA Gi\&Mi.         & 0.04      & 33.16     & 38.08     & \bf33.76$^*$     & \bf20.33$^*$     & 40.06     & 13.19     & \bf10.30     & \bf23.62$^*$     \\ 
Pal $\ell$=0*$\ell$=1  & 1.71      & \bf35.77$^*$     & \bf41.58     & 25.14     & 9.50      & 38.92     & \bf18.41     & 5.61      & 22.08     \\ 
Pal \textit{inter}+\textit{intra}      & \bf24.84$^*$     & 29.70     & 14.04     & 1.64      & 3.45      & 14.84     & 2.13      & 4.89      & 11.94     \\ 
\midrule
DBI*Mixup$^1$        & 0.00      & 25.86     & 32.05     & 31.79     & 15.92     & \bf43.99$^*$     & 12.59     & 9.24      & 21.43     \\

\bottomrule
\end{widetabular}
\end{center}
\end{table}

In the top part of Table \ref{tab:gen_pred_results}, we present CMI scores for measures that rely on only one form of Mixup-based measures, comparing Gi and Pal-scores to the one magnitude accuracy predictors from \cite{natekar2020representation}.
These results highlight that the Gi-score and Pal-score perform competitively in predicting generalization gap.
Note that some versions of our scores out-perform the mixup approaches used in the PGDL winning approach on the majority of tasks, and even substantially out-perform the DBI*Mixup approach on 5/8 tasks.
In addition, we believe that our scores provide a more robust measure for how well a model is able to learn invariances to certain transformations.
For example, the Mixup complexity score presented in \cite{natekar2020representation} simply takes a 0.5 interpolation of data points and calculates accuracy of a network on the this mixed up portion of the training set.
In contrast, our scores allow us to capture network performance on a spectrum of interpolations, thereby providing a more robust statistic for how invariant a network is to linear data interpolation.

\begin{figure}
     \centering
     \begin{subfigure}[b]{0.40\textwidth}
         \centering
         \includegraphics[width=\textwidth]{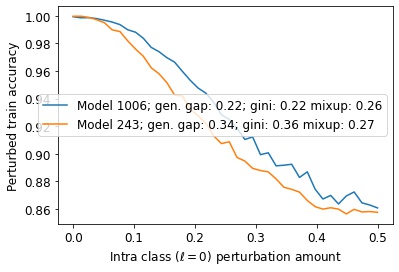}
         \caption{SVHN}
         \label{fig:prc_ex1}
     \end{subfigure}
     \begin{subfigure}[b]{0.40\textwidth}
         \centering
         \includegraphics[width=\textwidth]{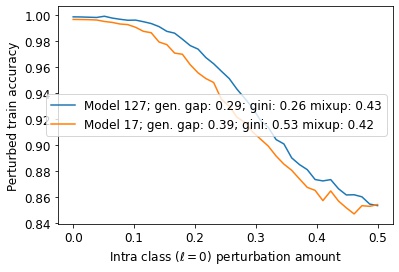}
         \caption{CIFAR-10 NiN}
         \label{fig:prc_ex2}
     \end{subfigure}
        \caption{Examples of PR Curves with normalized scores and gen. gaps for 2 different models showing different performance fall-off captured by Gi intra score, but mixup scores roughly the same.}
        \label{fig:pr_curve_compare}
\end{figure}

Indeed, we found examples of pairs of models that had significantly different generalization gap, but for which Mixup scores were roughly the same, and Gi-score was more reflective of the difference in generalization gap, and show a couple in Figure \ref{fig:pr_curve_compare} along with normalized scores and generalization gaps (normalized to be between 0 and 1 across all the models on the data).  Looking at the PR curves, we see the higher generalization gap model's PR curve deteriorating more quickly than the lower generalization gap model's PR curve, despite having roughly the same accuracy at $\alpha = 0.5$.  As a result, Mixup is not able to capture this difference between the PR curves and model sensitivity to perturbation, unlike our proposed framework and Gi-score which do capture this difference.

We note that for certain architectures different versions of the Gi and Pal-scores seem more useful.
Specifically, we observe that for NiN architectures Gi-scores for both inter and intra-class mixup on inputs are most informative, whereas for VGG architectures, Gi and Pal-scores on inter-class mixup from layer 1 representations are better predictors.
Future work will explore this pattern to better understand when each score is better suited for predicting generalization.
For fully convolutional architectures, both Gi and Pal-scores for intra-class mixup on inputs are good predictors of generalization gap, however, there is a performance drop compared to the PGDL competition winner on these network types, and Mixup accuracy of one magnitude alone works well on this dataset.
It seems for this particular architecture and/or dataset, the deterioration of accuracy at larger perturbations may be more important in predicting generalization.
However, even in this case we find that information from other parts of the perturbation response curve can still be useful, as will be described below in our combination scores.

In the bottom part of Table \ref{tab:gen_pred_results}, we show CMI scores for a few different ways of combining our Gi and Pal-scores (with different intra vs. inter class interpolation or at different layers) as well as Mixup (intra class interpolation with $\alpha = 0.5$).  Additionally we show the PGDL competition winner scores, which itself is a combination method taking the product of Mixup and DBI scores.
Scores and details for a larger systematic set of combinations of our scores are provided in Appendix \ref{app:complete_combo_res}.  

In the results shown here, ``PCA Gi\&Mi.'' corresponds to performing principal component analysis (PCA) on the pair of Gi intra $\ell$=0 scores and the Mixup scores per task and using the single score derived from projecting the pair of scores on the first principal component.
This corresponds to a setting in which we have a set of trained models for a dataset along with the training data and want to estimate their relative generalization performance (as opposed to being applicable to only a single model in isolation).
``Pal $\ell$=0*$\ell$=1'' corresponds to taking the product of Pal intra scores at layers 0 and 1, and ``Pal \textit{inter}+\textit{intra}'' refers to taking the average of Pal inter and Pal intra scores at layer 0.  

We found the best average CMI score (across all tasks) results from combining our Gi intra $\ell$=0 score (that summarizes the whole PR curve) with Mixup (that focuses on the end of the PR curve), via PCA.
As mentioned, this seems to suggest the general importance of focusing on multiple parts of the PR curve, paying particular attention to some regions -- in this case near the inversion perturbation amount $\alpha=0.5$ where the perturbed input is potentially farthest away from known examples but still in the same class region, since intra-class interpolation is used.

In general, we found a variety of different combination methods had higher average scores than the competition winner, and also different combinations of scores and methods gave better and sometimes significantly higher scores compared to most other methods for different tasks.
For example, simply taking the product of the Pal intra scores at layer 0 and layer 1 gives competitive scores and a higher score on average than the competition winner, and taking the average of Pal inter and Pal intra scores at layer 0 leads to a strikingly higher score on the CIFAR-10 VGG task than most methods (see Appendix \ref{app:complete_combo_res} for more examples).
This highlights the utility of having our different scores and the potential need to determine what measures work best for a particular architecture type and/or dataset.



\subsubsection{Generalization predictions: Timing and sensitivity analyses}\label{subsubsec:gen_pred_timing}
Our proposed PR curve framework is efficiently computed, as it consists of a simple forward pass of a model with computationally efficient perturbations on a subset of the training data.
Here we show results of experiments measuring PR curve generation time and CMI scores as a function of number of random batches of the training data, for different perturbation methods and layers for 2 datasets\footnote{PR curve generation takes the most time; score compute time is negligible in comparison (< 1 millisecond)}.
For timing analysis on the other tasks, see Appendix \ref{app:timing_and_sens}.
For each batch number experiment, we run 20 random runs and show the mean and standard deviation, to show the sensitivity to the size of the subsample used.
Each run is performed with 4 CPUs, 4 GB RAM, and 1 V100 GPU and batch size 128, submitted as resource-restricted jobs to a cluster.

\begin{figure}
     \centering
     \begin{subfigure}[b]{0.32\textwidth}
         \centering
         \includegraphics[width=\textwidth]{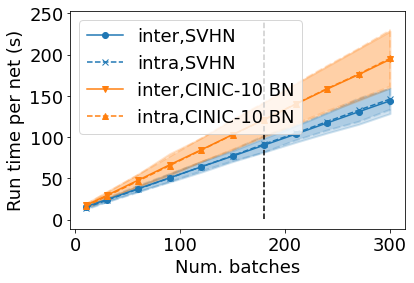}
         \caption{SVHN and CINIC-10 BN timing}
         \label{fig:cmi_main_timing}
     \end{subfigure}
     \hfill
     \begin{subfigure}[b]{0.32\textwidth}
         \centering
         \includegraphics[width=\textwidth]{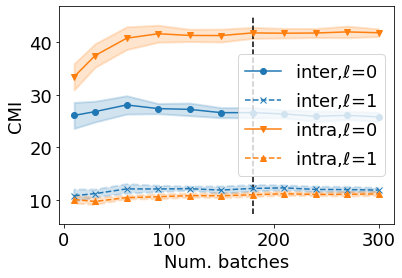}
         \caption{SVHN Gi-score CMI}
         \label{fig:cmi_main_svhn}
     \end{subfigure}
     \hfill
     \begin{subfigure}[b]{0.32\textwidth}
         \centering
         \includegraphics[width=\textwidth]{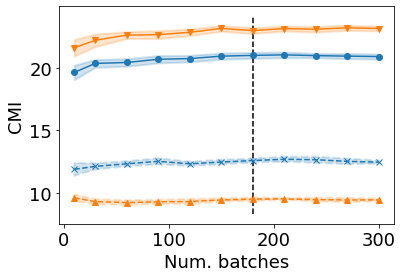}
         \caption{CINIC-10 BN Gi-score CMI}
         \label{fig:cmi_main_cinic10}
     \end{subfigure}
        \caption{Perturbation response curve generation run times (on input layer $\ell=0$) and Conditional Mutual Information score sensitivity results for 2 datasets. Mean and std. dev. over 20 runs vs. number of batches - 180 batches used in results table (dotted line).}
        \label{fig:time_and_sens}
\end{figure}

As seen in Figure \ref{fig:time_and_sens}, calculations on the number of batches used in our experiments (180) completes well within the PGDL competition 5 minute time limit (although the competition used 1 K80 GPU per score calculation as opposed to the 1 V100 GPU we used, it also used 4 CPUs as we did and 26GB RAM vs. 4 GB we used).
We could have even used more batches and still been within the time limit.
This illustrates that the proposed method is a good candidate for common use for evaluating network generalization ability, as it does not add significant computational burden.  Additionally the standard deviation of the CMI scores quickly becomes relatively small as the number of batches increases, indicating that these smaller subsamples of the training sets are sufficient to get consistent measure outputs, and that there is small variance in our reported average scores in Table \ref{tab:gen_pred_results}.

\subsection{Measuring invariance}\label{subsec:invariance}
To test whether our framework could accurately predict the degree to which a network has learned a given invariance, we create our own corpus of trained networks.
\subsubsection{Measuring invariance: Experimental setup}\label{subsubsec:invariance_exp}
We use two model architectures VGG and ResNet \cite{he2016deep} and train both architecture types on the  CIFAR-10 and SVHN datasets.
For the VGG networks, we use varying depths of 11, 13, 16, and 19 hidden layers with and without batch norm.
For the ResNet models, we use varying depths of 18, 34, and 50 hidden layers.
For all models, we train with batch sizes of either 1024, 2048, or 4096 and learning rates of either $1\rm{e}^{-4}$ or $1\rm{e}^{-5}.$
All models are trained with Adam optimization and a learning rate scheduler that reduced learning rate on training loss plateaus.

For each combination of model, dataset, batch norm application, batch size, and learning rate we train and test using four different types of parametric perturbations: (1) rotation, (2) horizontal translation, (3) vertical translation, and (4) color jittering.
The $\alpha_{\min}$ and $\alpha_{\max}$ for each perturbation are summarized in Table \ref{tab:perturbs}.
For rotation, the minimum and maximum perturbation values refer to degree of counter-clockwise rotation.
Note, that for SVHN, we use a smaller rotation range than for CIFAR-10, since SVHN contain digits, and we would therefore not want even the \textit{idealized} network to be invariant to the full $360^{\circ}$ range of rotations.
For horizontal translation, the minimum and maximum refer to the amount the image is shifted left or right relative to total image width, with negative values indicating a shift left and positive values indicating a shift right.
For vertical translation, the minimum and maximum refer to the amount the image is shifted up or down relative to total image height, with negative values indicating a shift up and positive values indicating a shift down.
For color jittering, the minimum and maximum refer to the amount by which brightness, contrast, saturation, and hue are adjusted.

\begin{table}[ht!]
  \caption{Perturbation minimum and maximum magnitudes by perturbation type and dataset.
  Minimum and maximums are displayed in each cell as an ordered pair.}
  \label{tab:perturbs}
  \centering
  \begin{tabular}{lcccc}
    \toprule
    & Rotation & Horizontal translation & Vertical translation & Color jittering \\
    \midrule
    CIFAR-10 & (-180, 179) & (-0.5, 0.5) & (-0.5, 0.5) & (-0.25, 0.25) \\
    SVHN     & (-90, 90)   & (-0.5, 0.5) & (-0.5, 0.5) & (-0.25, 0.25) \\
    \bottomrule
  \end{tabular}
\end{table}


For each perturbation, we train 3 versions of a given hyperparameter combination, one where no data augmentation is used in training, one where partial data augmentation is used in training, and one where full data augmentation is used.
Partial data augmentation means that training samples are randomly perturbed with up to 50\% of the perturbation minimum and maximum range.
Full data augmentation means that training samples are randomly perturbed within the full range.

For all training paradigms, the test set is randomly augmented with the full perturbation range and generalization gap are captured for each model.
Naturally, the models that are trained with full data augmentation have better accuracy on the test set, i.e., smaller generalization gap.

Finally, we calculate CMI scores as described in \cite{jiang2020neurips} for each architecture type, dataset, and perturbation combination, where the hyperparameters of depth, learning rate, and batch size (and batch norm for VGG networks) are used in the CMI calculation.

\subsubsection{Measuring invariance: Results}\label{subsubsec:invariance_res}
On each dataset, we calculate CMI scores for each of the two architecture types.
We repeat this calculation for each of the four perturbation types.
For each dataset, model, and perturbation type, we calculate the CMI score for Gi-scores as well as two baselines: (1) the model's accuracy on 10\% of the training data that is randomly augmented with the full perturbation range and (2) the mean accuracy for all $\alpha$'s in the PR curve, i.e., $\frac{1}{n_p}\sum_{\alpha_i=\alpha_{\min}}^{\alpha_{\max}} \mathcal{A}_{\alpha_i}^{(0)}.$
We chose these baselines as potentially good indicators of invariance to a specific perturbation, in order to see if our statistic provides added predictive capability.
The proposed complexity measures from the PGDL competition, such as Mixup from winning team \cite{natekar2020representation}, are less relevant in this context as baselines since they do not directly apply to measuring invariance of the perturbations that we test.
We only report results for models that were able to achieve at least 80\% accuracy on their respective training sets.

Finally, we note that for the perturbation ranges that we test in this section, the PR curves are not monotonically decreasing as they range from a large magnitude negative perturbation to a large magnitude positive perturbation, see for example Figure \ref{fig:pr_pcd_curve}, which shows the PR curve for a rotation perturbation.
Therefore, Pal-scores are omitted from this experiment because they are more applicable in cases where the PR curve is roughly monotonically decreasing.

The results of this experiment are presented in Table \ref{tab:perturbs_results}.
With higher CMI scores in almost all scenarios, these results highlight that the Gi-score is more informative than simply seeing how the model performs on a randomly augmented subset of the training data.
The results also demonstrate that extracting the Gi-score from PR curves is important, as this statistic is also better at predicting generalization gap than mean accuracy of the PR curve, although this naive alternative is also somewhat informative.

\begin{table}
  \caption{Conditional Mutual Information (CMI) scores for all dataset and model combinations across the four perturbations.
  Each dataset and model is presented in the columns, and the four perturbations are presented separately.
  For each perturbation, we report CMI for the Gi-score as well as two baselines: (1) the model's accuracy on 10\% of the training data that was randomly augmented with the full perturbation range and (2) the mean accuracy for all $\alpha$'s in the PR curve.
  Within each perturbation we only consider models that attain at least 80\% accuracy on their respective datasets, and we report the sample size of models for each setup along the row that indicates which perturbation we are examining.
  For each perturbation type, we bold the best CMI within a dataset-model combination.
  For all but two scenarios, the Gi-score provides the best predictive indicator of generalization gap.}
  \label{tab:perturbs_results}
  \centering
  \begin{tabular}{lcc|cc}
    \toprule
    \multicolumn{1}{c}{}
    &\multicolumn{2}{c|}{CIFAR-10}
    &\multicolumn{2}{c}{SVHN}\\
    \multicolumn{1}{c}{} & Resnet & VGG & Resnet & VGG\\
    \midrule\midrule
    \textit{Rotation} & ($n=34$) & ($n=93$) & ($n=49$) & ($n=142$)\\
    \midrule
    Acc. on augmented train subset & 27.99 & \bf16.99 & 47.24 & 42.97\\
    Mean acc. on PR curve & 27.61 & 15.61 & 48.14 & 44.05\\
    Gi-score & \bf41.54 & 15.29 & \bf54.11 & \bf46.11\\
    \midrule\midrule
    \textit{Horizontal translation} & ($n=36$) & ($n=112$) & ($n=50$) & ($n=143$)\\
    \midrule
    Acc. on augmented train subset & 41.79 & 33.48 & 29.49 & 24.20\\
    Mean acc. on PR curve & 45.03 & 33.00 & 29.88 & 24.28\\
    Gi-score & \bf50.07 & \bf34.31 & \bf34.56 & \bf25.94\\
    \midrule\midrule
    \textit{Vertical translation} & ($n=36$) & ($n=107$) & ($n=49$) & ($n=141$)\\
    \midrule
    Acc. on augmented train subset & 26.79 & 35.68 & 51.88 & 50.98\\
    Mean acc. on PR curve & 26.55 & 37.33 & 52.39 & 52.22\\
    Gi-score & \bf34.85 & \bf39.07 & \bf59.02 & \bf52.83\\
    \midrule\midrule
    \textit{Color-jittering} & ($n=44$) & ($n=130$) & ($n=49$) & ($n=143$)\\
    \midrule
    Acc. on augmented train subset & 35.77 & 28.44 & 43.08 & \bf30.07\\
    Mean acc. on PR curve & 39.12 & 29.37 & 43.26 & 28.67\\
    Gi-score & \bf44.63 & \bf30.79 & \bf50.08 & 29.45\\
    \bottomrule
  \end{tabular}
\end{table}




\section{Conclusion}\label{sec:conclusion}

In this work, we introduced a general framework, consisting of computing perturbation response (PR) curves, and two novel statistics, inspired by income inequality metrics, which effectively predict network generalization.
The Gi and Pal-scores are intuitive and computationally efficient, requiring only several forward passes through a subset of the training data.

Calculating these statistics on the corpus of data made available in \cite{jiang2020neurips} showed that our scores applied to linear interpolation between data points have strong performance in predicting a network's generalization gap.
In addition to this predictive capability of generalization, we demonstrated that our framework and the Gi-score provide a useful criterion for evaluating networks and selecting which architecture or hyperparameter configuration is most invariant to a given transformation.

Because they rely on comparison with an idealized network that remains unaffected at all magnitudes of a perturbation, our scores are more informative than other baselines that rely on examining how a trained network responds to perturbed data.  Specifically, it is interesting to note that our scores provide more informative measures than simply averaging the PR curves or taking the accuracy for random perturbation magnitudes -- suggesting our scores may better capture varying effects of different perturbation amounts (perhaps related to discoveries presented in \cite{yin2019fourier} for augmenting training).
Future work will explore this hypothesis further.
Future work will also demonstrate the usefulness of our statistics on other parametric perturbations and will compare the Gi and Pal-scores of architectures that are known to be more invariant to certain perturbations, e.g., those described in \cite{bronstein2021geometric}, to those for models without the inductive biases that contribute to invariance.


Although we provide strong empirical evidence that our framework and Gi and Pal-scores are distinctly informative of network generalization and invariance, currently our approach lacks  theoretical underpinning of the performance obtained by these scores.
Future work includes deriving theoretical analyses detailing under which conditions our scores correctly identify the best network and correctly rank networks based on generalization, as well as consistency guarantees. 
Additionally, we plan to evaluate our framework on other tasks and modalities, such as language models and time series.  We also plan to explore the possibility of regularization approaches based on our scores. 


\bibliographystyle{IEEEtran}
\bibliography{references}

\newpage
\appendix

\section{Appendix}\label{appendix}
\subsection{Assets}\label{app:assets}
For the generalization prediction experiments, we use the data provided by the PGDL competition organizers \cite{jiang2020neurips}, which is available for download here: \href{https://github.com/google-research/google-research/tree/master/pgdl\#wheres-the-data}{https://github.com/google-research/google-research/tree/master/pgdl\#wheres-the-data} and is released under the Apache 2.0 license, and corresponding Tensorflow \cite{tensorflow2015-whitepaper} code for loading models, and wrote our perturbation code using Tensorflow as well.

For the measuring invariance experiments, we use the CIFAR-10 \cite{krizhevsky2009learning} dataset, released under the MIT license, and SVHN dataset \cite{netzer2011reading}, which does not have a license listed with the dataset.
Both datasets were downloaded and split into training and test sets by the \texttt{torchvision} module of the PyTorch \cite{paszke2019pytorch} library.

The packages used in this work and their respective licenses are listed below:
\begin{enumerate}
    \item PGDL Competition Starter Kit \cite{jiang2020neurips}; Apache 2.0
    \item PyTorch \cite{paszke2019pytorch}; BSD
    \item PyTorch Lightning \cite{falcon2019pytorch}; Apache 2.0
    \item Tensorflow \cite{tensorflow2015-whitepaper}; Apache 2.0
\end{enumerate}


\subsection{Calculating Conditional Mutual Information scores}\label{app:cmi}
Given it's importance to our analyses, we reproduce the calculation for Conditional Mutual Information (CMI) scores presented in \cite{jiang2020neurips} here.

Generalization gap is defined as:
$$g(f, \mathcal{D}) = \sum_{x, y \in \mathcal{D}_{train}}\one(\max_{i \in [k]}f(x)[i] = y) \ / \ |\mathcal{D}_{train}| - \sum_{x, y \in \mathcal{D}_{test}}\one(\max_{i \in [k]}f(x)[i] = y) \ / \ |\mathcal{D}_{test}|$$

The goal of the PGDL competition was to find a complexity measure $\mu$ such that:
$$\mathrm{sgn}(\mu(f, \mathcal{D}_{train}), \mu(f', \mathcal{D}_{train})) = \mathrm{sgn}(g(f, \mathcal{D}), g(f', \mathcal{D}))$$

Let
$$V_g(f, f') = \mathrm{sgn}(g(f, \mathcal{D}), g(f', \mathcal{D}))$$
and 
$$V_{\mu}(f, f') = \mathrm{sgn}(\mu(f, \mathcal{D}_{train}), \mu(f', \mathcal{D}_{train}))$$

Now, we use $\mathcal{O}$ to denote the set of hyperparameters.
For example, in Section \ref{subsec:invariance}, for Resnet models we use $\mathcal{O} = \{\mathrm{depth}, \mathrm{learning \ rate}, \mathrm{batch \ size}\}$ so that $|\mathcal{O}| = 3.$
Models can be separated into groups based on their specific value for each hyperparameter.
Each group is denoted as $\mathcal{O}_k$, i.e., the set of models that have hyperparameter configuration $k$.
For each hyperparameter, $\Theta_i \in \mathcal{O}$, $|\Theta_i|$ is the number of possible values that parameter can take on, so that for example, $\Theta_i =$ Resnet depths of 18, 34, 50, we have $|\Theta_i| = 3$.
Thus the total number of groups is $\prod_{\Theta_i \in \mathcal{O}}{|\Theta_i|}$.

If we treat $V_g$ and $V_{\mu}$ as Bernoulli random variables, then we can calculate the probabilities:
$$p(V_g | \mathcal{O}_k), \ \ p(V_{\mu} | \mathcal{O}_k), \ \ p(V_g, V_{\mu} | \mathcal{O}_k)$$
where the probabilities are calculated by counting over models in each group $\mathcal{O}_k.$

With these probabilities, we can define mutual information between $V_g$ and $V_\mu$:
$$\mathcal{I}(V_g, V_\mu | \mathcal{O}_k) = \sum_{V_g}\sum_{V_\mu}{p(V_g, V_{\mu} | \mathcal{O}_k)}\log \Big( \frac{p(V_\mu, V_g | \mathcal{O})}{p(V_g|\mathcal{O}_k)p(V_{\mu}|\mathcal{O}_k)} \Big)$$

Each $\mathcal{O}_k$ occurs with the same probability $p_c = 1 /\prod_{\Theta_i \in \mathcal{O}}{|\Theta_i|}.$
Thus, using the same notation abuse as in \cite{jiang2020neurips}, we have mutual information between $V_\mu$ and $V_g$ conditioned on the values of $\mathcal{O}$:
$$\mathcal{I}(V_g, V_\mu | \mathcal{O}) = \sum_{\mathcal{O}_k}{p_c\mathcal{I}(V_g, V_\mu | \mathcal{O}_k)}$$

To get values between 0 and 1, we can normalize this conditional mutual information by conditional entropy of generalization, which is defined as:
$$\mathcal{H}(V_g | \mathcal{O}) = \sum_{\mathcal{O}_k}{p_c}\sum_{V_g}{\log(p(V_g | \mathcal{O}_k))}$$

So we now have normalized conditional mutual information:
$$\hat{\mathcal{I}}(V_g, V_\mu | \mathcal{O}) = \frac{\mathcal{I}(V_g, V_\mu | \mathcal{O})}{\mathcal{H}(V_g | \mathcal{O})}$$

Finally, the CMI score used in PGDL and presented in the results sections in this work is defined as:
$$\mathrm{CMI}(\mu) = \min_{\mathcal{O}}\hat{\mathcal{I}}(V_g, V_\mu | \mathcal{O})$$


\subsection{Algorithm for generating Perturbation Response Curves}
\label{app:alg_prc}

Here we provide the detailed algorithm pseudo-code for generating perturbation response curves, described in the main paper.

\begin{algorithm}
\DontPrintSemicolon
  \SetKwInput{KwInputs}{Inputs}                
  \SetKwInput{KwOutput}{Output}              
  \KwInputs{Trained model $f$; Dataset $\mathcal{D}$; Perturbation $\mathcal{T}_\alpha$; Min perturbation magnitude $\alpha_{\min}$; Max perturbation magnitude $\alpha_{\max}$; Number of perturbation magnitudes to measure $n_p$; Layer at which to apply the perturbation $\ell$; number of batches to sample $n_b$; batch size $b_s$}
  \KwOutput{PR Curve: Arrays of regularly spaced perturbation magnitudes ranging from $\alpha_{\min}$ to $\alpha_{\max}$ of length $n_p$ [$\alpha_{\min}$, $\alpha_{\max}$][$n_p$] and accuracy array at each perturbation magnitude of length $n_p$ $\mathcal{A}_\alpha$[$n_p$]}
  \For{$i \gets 0$ to $n_p-1$}{
        $\alpha_i$ $\leftarrow$ [$\alpha_{\min}$, $\alpha_{\max}$][$i$]\\
        Shuffle $\mathcal{D}$\;
        \For{$k \gets 0$ to $n_b-1$}{
            $\mathcal{D}_{sample} \gets  \mathcal{D}[k b_s:(k+1) b_s]$  \tcp{batch $k$ of $\mathcal{D}$}    
            $\mathcal{A}_{\alpha_i}^{(\ell)}$[k] $\gets $ batch accuracy under perturbation $\mathcal{T}_{\alpha_i}$ (Equation \ref{eq:perturb_acc})
        }
        $\mathcal{A}_\alpha$[i] $\leftarrow$ $\sum_k \mathcal{A}_{\alpha_i}^{(\ell)}$[k]$/n_b$ \\
        
  }
\caption{Building Perturbation Response (PR) Curve}
\label{alg:pr_curve}
\end{algorithm}

\subsection{Algorithm for Gi-scores} \label{app:alg_gini}


Here we provide the detailed algorithm described in the main paper for computing the Gi-scores, shown in Algorithm \ref{alg:gi_score}.

\begin{algorithm}
\DontPrintSemicolon
  \SetKwInput{KwInputs}{Inputs}                
  \SetKwInput{KwOutput}{Output}              
  \KwInputs{Arrays of perturbation magnitude $\alpha$[$n$] and accuracy $\mathcal{A}_\alpha$[$n$]}
  \KwOutput{Gi-score $gi$}
  $a_t[0] \gets 0$ \tcp{initialize 1st element of trapezoidal areas array with 0}
  \For{$i \gets 0$ to $n-2$}{
    $a_t[i+1] \gets 0.5 (\alpha[i+1] - \alpha[i])(\mathcal{A}_\alpha[i] + \mathcal{A}_\alpha[i+1])$
  }
  \For{$i \gets 1$ to $n-1$}{
    $a_t[i] \gets a_t[i] + a_t[i-1]$. \tcp{cumulative sum}
  } 
  $d[i] =  \alpha[i] - a_t[i] , \forall i$\\
  $gi=0$ \;
  \For{$i \gets 0$ to $n-2$}{
    $gi \gets gi + 0.5 (\alpha[i+1] - \alpha[i])(d[i] + d[i+1])$
  }
  $gi \gets gi / (0.5 \alpha[n-1]^2)$ \tcp{Divide by area under line of equality}
  \Return  $gi$ \;
\caption{Gi-Score computation given PR Curve for a model}
\label{alg:gi_score}
\end{algorithm}

\subsection{Algorithm for efficiently computing interpolation PR curves}\label{app:alg_mixup}

Here we give more details about how we compute the accuracy under interpolation per batch -- for the intra class interpolation.
In order to do this efficiently, we compute accuracy per randomly sampled batches with simple operations that can be encoded in a computational graph (e.g., Tensorflow), and then compute the mean accuracy from multiple batch accuracies. 
Inter-class interpolation is performed similarly.
In each case, we split the batch into pairs efficiently, by random pairing for inter-class interpolation since probability of being from different classes is high, and by sorting by class label and then interleaving the sorted entries to get the pairs, since it is most likely to get pairs of the same class this way.
In both cases we throw out any pairs from the batch that do not match (any pairs having the same class for inter-class interpolation and any pairs having different class for intra-class interpolation).
We then compute the accuracy for the batch, and keep track of the effective batch size, to update the total accuracy across all batches.  

\begin{algorithm}[htb]
\DontPrintSemicolon
  \SetKwInput{KwInputs}{Inputs}                
  \SetKwInput{KwOutput}{Output}              
  \KwInputs{Sampled batch of data for layer $\ell$ to perturb $x^{(\ell)},y$ of size $n$ (assume here $n$ is an even number to simplify description), perturbation magnitude $\alpha \in [0,1]$,  and network function $f_{\ell+1}$}
  \KwOutput{Accuracy for batch sample $\mathcal{A}_\alpha^{(\ell)}$ }
  Sort $x^{(\ell)},y$ by class labels $y$ \;
  \tcp{Assign every other element, starting with first element}
  $x^{(\ell)}_1 \gets x^{(\ell)}[::2]$ \; 
  $y_1 \gets y[::2]$ \;
  \tcp{Assign every other element, starting with second element}
  $x^{(\ell)}_2 \gets x^{(\ell)}[1::2]$ \; 
  $y_2 \gets y[1::2]$ \;
  \tcp{Each entry at index $i$ in $x^{(\ell)}_1$ and $x^{(\ell)}_2$, and $y_1$ and $y_2$ form pairs}
  Drop any index $i$ in  $x^{(\ell)}_1$, $x^{(\ell)}_2$, $y_1$, and $y_2$ where $y_1[i] \neq y_2[i]$ \;
  
  $x^{(\ell)}_p \gets (1-\alpha)x^{(\ell)}_1 + \alpha x^{(\ell)}_2$  \tcp{Get interpolated point} \;
   
  \tcp{Compute mean accuracy for interpolated points}
  $\mathcal{A}_\alpha^{(\ell)} = \sum_{j}\one(\max_{i \in [k]}f_{\ell + 1}(x^{(\ell)}_{p}[j])[i] = y_{1}[j]) \ / \ |y_1| $ \;

  \Return  $\mathcal{A}_\alpha^{(\ell)}$ \;
\caption{Efficient computation of intra-class interpolation for a batch}
\label{alg:intra_interp_batch_acc}
\end{algorithm}

\subsection{Algorithm for computing Pal-scores}\label{app:alg_pal}
In this section, we present the pseudocode for calculating the Pal-score, in Algorithm \ref{alg:pal_score}, which is similar to the Gi-score calculation, but with a focus on certain areas of the PCD curve.

\begin{algorithm}[htb]
\DontPrintSemicolon
  \SetKwInput{KwInputs}{Inputs}                
  \SetKwInput{KwOutput}{Output}              
  \KwInputs{Arrays of perturbation magnitude $\alpha$[$n$] and accuracy $\mathcal{A}_\alpha$[$n$]}
  \KwOutput{Pal-score $pal$}
  $a_t[0] \gets 0$ \tcp{initialize 1st element of trapezoidal areas array with 0}
  \For{$i \gets 0$ to $n-2$}{
    $a_t[i+1] \gets 0.5 (\alpha[i+1] - \alpha[i])(\mathcal{A}_\alpha[i] + \mathcal{A}_\alpha[i+1])$
  }
  $top\_idx \gets$ index of 60\% of $\alpha[n]$ \;
  $bottom\_idx \gets$ index of 10\% of $\alpha[n]$ \;
  $pal \gets a_t[top\_idx] / a_t[bottom\_idx]$ \;
  
  \Return  $pal$ \;
\caption{Pal-Score computation given PR Curve for a model}
\label{alg:pal_score}
\end{algorithm}

\subsection{Additional sensitivity analysis}\label{app:timing_and_sens}
\begin{figure}[htb]
     \centering
     \begin{subfigure}[b]{0.32\textwidth}
         \centering
         \includegraphics[width=\textwidth]{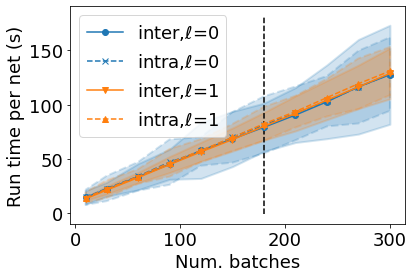}
         \caption{CIFAR-10 NiN timing}
     \end{subfigure}
     \hfill
     \begin{subfigure}[b]{0.32\textwidth}
         \centering
         \includegraphics[width=\textwidth]{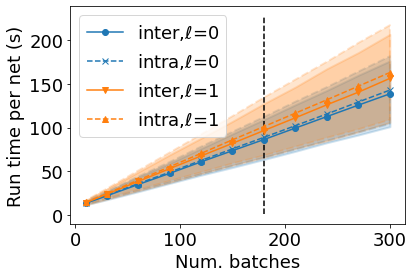}
         \caption{CIFAR-10 VGG timing}
     \end{subfigure}
     \hfill
     \begin{subfigure}[b]{0.32\textwidth}
         \centering
         \includegraphics[width=\textwidth]{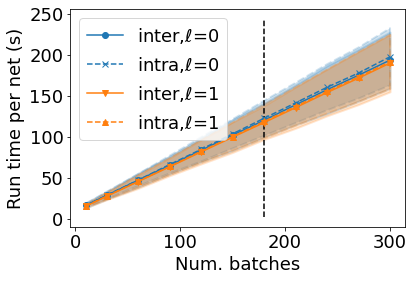}
         \caption{CINIC-10 BN timing}
     \end{subfigure}
     \hfill
     \begin{subfigure}[b]{0.32\textwidth}
         \centering
         \includegraphics[width=\textwidth]{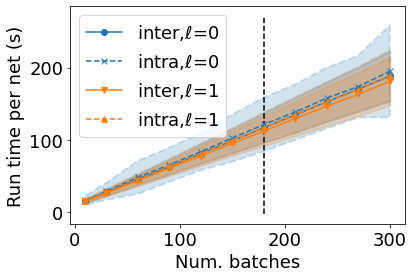}
         \caption{CINIC-10 No BN timing}
     \end{subfigure}
     \hfill
     \begin{subfigure}[b]{0.32\textwidth}
         \centering
         \includegraphics[width=\textwidth]{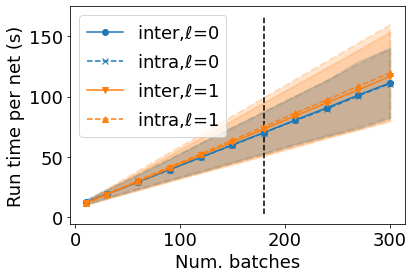}
         \caption{Fashion MNIST timing}
     \end{subfigure}
     \hfill
     \begin{subfigure}[b]{0.32\textwidth}
         \centering
         \includegraphics[width=\textwidth]{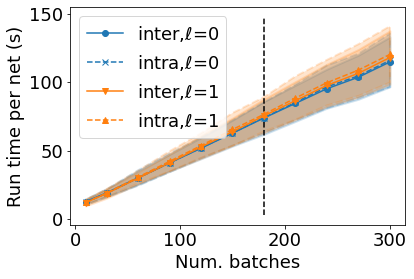}
         \caption{Oxford Flowers timing}
     \end{subfigure}
     \hfill
     \begin{subfigure}[b]{0.32\textwidth}
         \centering
         \includegraphics[width=\textwidth]{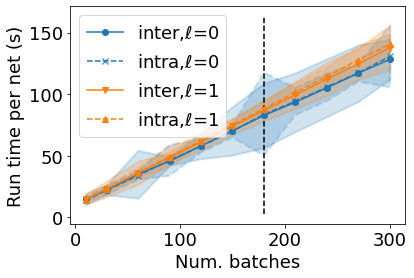}
         \caption{Oxford Pets timing}
     \end{subfigure}
     \begin{subfigure}[b]{0.32\textwidth}
         \centering
         \includegraphics[width=\textwidth]{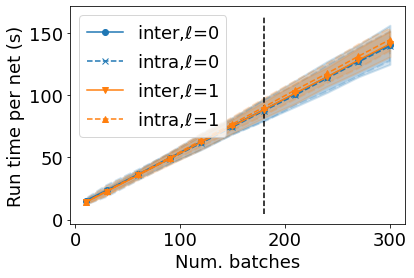}
         \caption{SVHN timing}
     \end{subfigure}
     \hfill
        \caption{Perturbation response curve generation run times for each datasets on input layer and layer 1 ($\ell=0$ and $\ell=1$). Mean and std. dev. over 20 runs vs. number of batches -- 180 batches used in results table (dotted line) }
        \label{fig:timing_more}
\end{figure}

\begin{figure}[htb]
     \centering
     \begin{subfigure}[b]{0.32\textwidth}
         \centering
         \includegraphics[width=\textwidth]{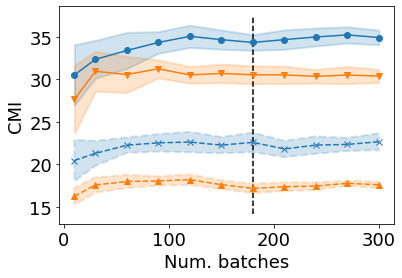}
         \caption{CIFAR-10 NiN CMI}
     \end{subfigure}
     \hfill
     \begin{subfigure}[b]{0.32\textwidth}
         \centering
         \includegraphics[width=\textwidth]{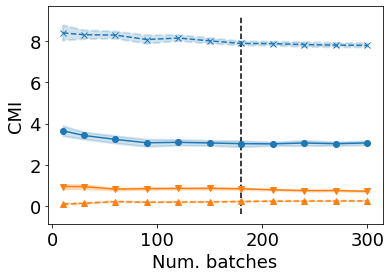}
         \caption{CIFAR-10 VGG CMI}
     \end{subfigure}
     \hfill
     \begin{subfigure}[b]{0.32\textwidth}
         \centering
         \includegraphics[width=\textwidth]{figures/cmi/cmi_per_batch_CINIC10_BN_legFalse.png}
         \caption{CINIC-10 BN CMI}
     \end{subfigure}
     \hfill
     \begin{subfigure}[b]{0.32\textwidth}
         \centering
         \includegraphics[width=\textwidth]{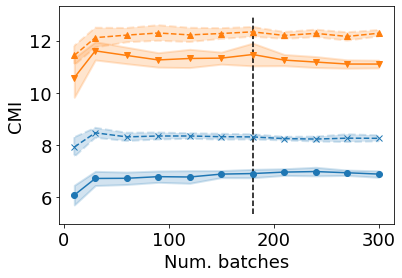}
         \caption{CINIC-10 No BN CMI}
     \end{subfigure}
     \hfill
     \begin{subfigure}[b]{0.32\textwidth}
         \centering
         \includegraphics[width=\textwidth]{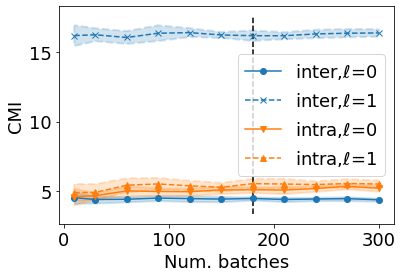}
         \caption{Fashion MNIST CMI}
     \end{subfigure}
     \hfill
     \begin{subfigure}[b]{0.32\textwidth}
         \centering
         \includegraphics[width=\textwidth]{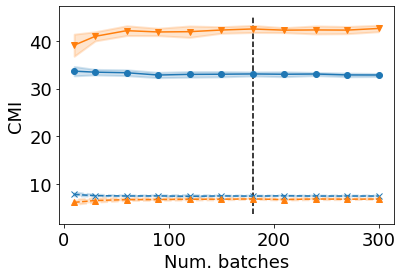}
         \caption{Oxford Flowers CMI}
     \end{subfigure}
     \hfill
     \begin{subfigure}[b]{0.32\textwidth}
         \centering
         \includegraphics[width=\textwidth]{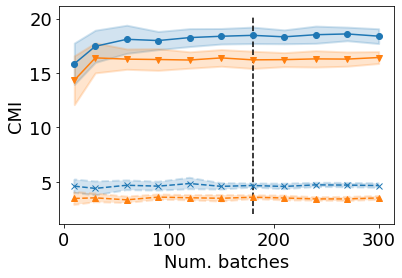}
         \caption{Oxford Pets CMI}
     \end{subfigure}
     \begin{subfigure}[b]{0.32\textwidth}
         \centering
         \includegraphics[width=\textwidth]{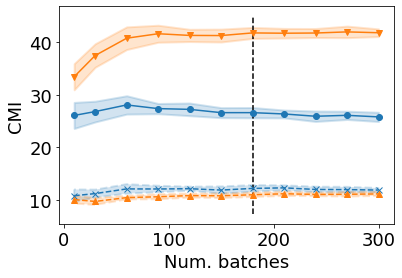}
         \caption{SVHN CMI}
     \end{subfigure}
     \hfill
        \caption{Conditional Mutual Information (CMI) score sensitivity results for all datasets on input layer and layer 1 ($\ell=0$ and $\ell=1$). Mean and std. dev. over 20 runs vs. number of batches -- 180 batches used in results table (dotted line) }
        \label{fig:cmi_more}
\end{figure}

Here we show the timing (Figure \ref{fig:timing_more}) and Gi-score CMI (Figure \ref{fig:cmi_more}) sensitivity results for all datasets for both intra and inter-class perturbations, and for both input ($\ell = 0$) and first hidden layer ($\ell = 1$).
This analysis is averaged over 20 different runs (each having different random sampling of the full data set per batch) -- with mean and standard deviation shown.

The first set of plots shows the timing results for all datasets (Figure \ref{fig:timing_more}).
We see that for plots, even those with larger number of batches, the time taken to compute our scores is well under the competition time limit -- and the longest time dataset (CINIC-10) was already reported in the main paper.

Figure \ref{fig:cmi_more} shows the Gi-score CMI sensitivity to number of batches (mean and std. dev.) for all datasets, for both perturb types (inter and intra-class) and for both layers (input - layer 0, and layer 1).
This confirms the stability of the scores with sufficient sub-sample size (number of batches) and that the number of batches was chosen to be large enough to ensure stable scores in our reported results.

\subsection{Measuring generalization: Complete GI and Pal Score combination results}\label{app:complete_combo_res}
In this section, we show results for a complete systematic set of combinations of our different measures (Gi and Pal-scores with different intra and inter-class interpolation, and at different layers, 0 and 1) as well as Mixup (intra-class interpolation with $\alpha = 0.5$) in select cases, for different score combination approaches.
This is not an exhaustive set of combinations, but fairly broad coverage of possible combinations, using only relatively simple approaches.

We also include two types of combinations that require a set of different models / complete set of models for a task before the final score can be derived to estimate generalization: principle component analysis (PCA) and rank-based combinations.   This could correspond to the setting in which we have a set of trained models for a dataset along with the training data and want to estimate their relative generalization performance (as opposed to being applicable to only a single model).  Details of how the combined scores are computed with these two approaches are given in the following sections along with the other combination methods that do not have this constraint.

The types of combinations we used are the following, which are described in detail in the subsequent sections, along with  corresponding results tables:
\begin{itemize}
    \item PCA and NPCA
    \item AVG
    \item PROD
    \item PROD+AVG
    \item AVG RANK
\end{itemize}

Note, for any combination of Gi and Pal-scores, we use the negative of the Pal-scores, since Gi and Pal-scores are oppositely correlated (anti-correlated).
Therefore, in order to support all forms of combining the scores, we use the negative of Pal-score when combining with Gi-score, as some combination methods would not otherwise work as expected if they are oppositely correlated, e.g., taking the average of the two scores.

It is also interesting to note that for any type of combination, there are some combinations that work well on average across tasks -- i.e., better than the competition winner average score across tasks, but different combinations of different methods work best on different tasks.

\subsubsection{PCA and NPCA (Tables \ref{tab:combo_pca_gen_pred_results} and \ref{tab:combo_norm_pca_gen_pred_results} resp.)}
Given the full set of models and scores for each task, to get the combined score we perform principal component analysis (PCA) on the set of scores being combined and project the multiple scores on the first principal component to get a single combination score for each model.
The intuition is to combine the scores by capturing the principal direction of variation amongst the two or more score dimensions, for each task.
We combine pairs of various Gi and Pal=scores, as well as Gi-score with Mixup, and also include combinations of all Gi and Pal-scores (``PCA all Gi \& Pal''), all Gi-scores (``PCA all Gi'') and all Pal-scores (``PCA all Pal'').
We also include in the second table the results using PCA after normalizing / standardizing the individual scores (``NPCA'') across the models in each task, i.e., subtracting the mean and dividing by standard deviation.  
    
As shown in the main paper, the combination that gives the best average score comes from combining Gi intra at level 0 with Mixup.  Various combinations do better at different tasks.
The fact that combining the Gi-score and Mixup at level 0, both with intra-class interpolation, does best on average may suggest that it is important to pay attention to both how quickly the PR curve falls off for varying amounts of perturbation as well as how much it falls off near the inversion point of $\alpha=0.5$, where it is farthest away from any training example. 
This may encourage developing alternate scores from the PR curves as well.
    
It is also interesting to note that combining more than 2 methods in this way, such as all Gi and Pal-scores, does not give the best results or better scores in general than combining pairs.
This may be because with more scores included, the most degree of variation can come from differences in the scores that are not connected to generalization, and choosing only one principal component dimension limits what is captured from the multiple scores.
Without supervisory information (i.e., test errors) the PCA components found may not line up with generalization in general. 
In general, combining multiple scores in an unsupervised way can be a challenging task, so another possibility for future work is looking for ways to combine scores given some labeled training data (in the form of models with generalization gaps known for given datasets) such that these combinations also work for unseen tasks (datasets and models).
This echoes the discussion above, as learning how to combine the scores from different areas of the PR curve may give the best results, as might be suggested by seeing good results with the combination of Gi intra $\ell$0 and Mixup.

\subsubsection{AVG (Table \ref{tab:combo_avg_gen_pred_results})}
For these set of results, we simply average the two scores together.
In this case, a simple average of the Gi intra score at level 0 and the Pal intra score at level 0 (again taking the negative of the latter since it is anti-correlated with Gi-score) yields the best average score across tasks.  
    
It is also interesting to note that certain combinations of the scores across layers and interpolation types give significantly better scores compared to most other methods for certain specific tasks.
For instance, combining Gi intra $\ell$0 and Gi inter $\ell$1 gives a higher score on Fashion MNIST, and combining Pal intra $\ell$0 and Pal inter $\ell$0 gives a strikingly higher score on the CIFAR-10 VGG task compared to most other methods.
    
\subsubsection{PROD (Table \ref{tab:combo_prod_gen_pred_results})}
    For this combination approach, we take the product of the two scores being combined.  In this case the product of Pal intra $\ell$0 and Pal inter $\ell$0 gave the best results on average.
    
\subsubsection{PROD+AVG (Table \ref{tab:combo_prod_avg_gen_pred_results})} 
This combination approach combines the previous two. 
The product of the two scores being combined is added to their average, to obtain the final single score.
    
\subsubsection{AVG RANK (Table \ref{tab:combo_avg_rank_gen_pred_results})}
For this and the next combination method, raw scores for each individual score are first transformed to ranks per task (model and dataset combination), by ranking the scores from smallest to largest, and assigning the smallest rank a score of 1 and the largest a score of the number of models in the task.  For AVG RANK, the average is taken between the rank-transformed scores of two different scores. 
Again, if a Gi-score is combined with a Pal-score, the negative of the Pal-score is used, since Gi and Pal-scores are anti-correlated.
    
For this AVG RANK approach, Gi intra $\ell$0 combined with Pal intra $\ell$0 gave the best average result, suggesting some useful and different information is captured by the different score approaches, so combining them can be useful.
    

\begin{table}[ht!]
\small
\caption{Comparison of Conditional Mutual Information scores for various complexity measures across tasks.
We present combinations of multiple of our measures using PCA per task,  and include the PGDL competition winner scores at the bottom.
The highest score within each task is bolded.
In the CINIC-10 columns, `bn' stands batch-norm.
}
\label{tab:combo_pca_gen_pred_results}
\begin{center}
\begin{widetabular}{\textwidth}{l | c c | c | c c | c | c | c | c}
\toprule 
\multicolumn{1}{c|}{}
&\multicolumn{2}{c|}{CIFAR-10}
&\multicolumn{1}{c|}{SVHN}
&\multicolumn{2}{c|}{CINIC-10}
&\multicolumn{1}{c|}{\makecell{Oxford \\ Flowers}}
&\multicolumn{1}{c|}{\makecell{Oxford \\Pets}}
&\multicolumn{1}{c|}{\makecell{Fashion \\ MNIST}}
&\makecell[t]{\textit{All} \\ \textit{Avg}}
\\\cline{2-9}
\multicolumn{1}{c|}{}
&\textit{VGG}
&\textit{NiN}
&\textit{NiN}
&\makecell{\textit{Conv}\\\textit{w/bn}}
&\makecell{\textit{Conv}\\}
&\textit{NiN}
&\textit{NiN}
&\textit{VGG}
\\ \midrule
PCA all Gi \& Pal                             & 5.08         & 28.24        & 18.22        & 19.80        & 10.71        & 20.10        & 5.03         & 7.36         & 14.32        \\ 
PCA all Gi                                   & 4.98         & 25.49        & 17.89        & 17.42        & 9.84         & 20.30        & 6.56         & 6.25         & 13.59        \\ 
PCA all Pal                                  & 5.08         & 28.24        & 18.22        & 19.80        & 10.71        & 20.10        & 5.03         & 7.36         & 14.32        \\ 
PCA Gi intra $\ell$0 \& Gi intra $\ell$1      & 0.11         & 25.51        & 19.06        & 16.18        & 12.14        & 16.02        & 6.37         & 5.38         & 12.60        \\ 
PCA Gi intra $\ell$0 \& Gi inter $\ell$0      & 2.02         & 33.33        & 35.06        & 22.24        & 8.95         & 37.70        & 17.05        & 4.55         & 20.11        \\ 
PCA Gi intra $\ell$0 \& Gi inter $\ell$1      & 6.24         & 24.91        & 21.83        & 17.84        & 9.64         & 12.32        & 6.20         & 16.68        & 14.46        \\ 
PCA Gi intra $\ell$0 \& Pal intra $\ell$0     & 0.88         & 35.85        & \bf43.51     & 26.72        & 12.89        & \bf43.44     & 18.00        & 7.02         & 23.54        \\ 
PCA Gi intra $\ell$0 \& Pal intra $\ell$1     & 0.25         & 15.62        & 12.35        & 9.89         & 13.07        & 7.02         & 5.23         & 8.32         & 8.97         \\ 
PCA Gi intra $\ell$0 \& Pal inter $\ell$0     & 3.33         & \bf37.22     & 32.37        & 23.90        & 7.69         & 29.62        & 17.14        & 5.35         & 19.58        \\ 
PCA Gi intra $\ell$0 \& Mix intra $\ell$0     & 0.04         & 33.16        & 38.08        & \bf33.76     & \bf20.33     & 40.06        & 13.19        & 10.30        & \bf23.62     \\ 
PCA Gi intra $\ell$1 \& Gi inter $\ell$0      & 7.77         & 26.67        & 16.41        & 16.31        & 9.76         & 28.29        & 6.30         & 4.52         & 14.50        \\ 
PCA Gi intra $\ell$1 \& Mix intra $\ell$0     & 0.01         & 32.38        & 32.71        & 32.97        & 19.92        & 39.59        & 13.96        & 10.78        & 22.79        \\ 
PCA Gi inter $\ell$0 \& Gi inter $\ell$1      & \bf8.05      & 25.93        & 19.92        & 16.92        & 8.30         & 20.41        & 6.46         & 5.86         & 13.98        \\ 
PCA Gi inter $\ell$0 \& Mix intra $\ell$0     & 0.49         & 33.52        & 38.61        & 33.72        & 18.94        & 40.96        & 13.63        & 5.36         & 23.16        \\ 
PCA Gi inter $\ell$1 \& Mix intra $\ell$0     & 0.10         & 30.08        & 34.92        & 33.46        & 18.50        & 36.20        & 13.12        & 16.95        & 22.91        \\ 
PCA Pal intra $\ell$0 \& Pal inter $\ell$0    & 2.36         & 36.18        & 42.26        & 25.14        & 9.82         & 38.43        & \bf18.35     & 5.50         & 22.26        \\ 
PCA Pal intra $\ell$0 \& Pal inter $\ell$1    & 6.79         & 23.88        & 21.05        & 21.04        & 10.60        & 13.82        & 4.87         & \bf17.12     & 14.90        \\ 
\hline
\hline \hline
\textit{DBI*Mixup}                       & \it0.00      & \it25.86     & \it32.05     & \it31.79     & \it15.92     & \it43.99     & \it12.59     & \it9.24      & \it21.43     \\ 

\bottomrule
\end{widetabular}
\end{center}
\end{table}

\begin{table}[ht!]
\small
\caption{Comparison of Conditional Mutual Information scores for various complexity measures across tasks.
We present combinations of multiple of our measures using PCA per task after normalizing / standardizing each score per task (subtracting mean and dividing by std. dev. across the task),  and include the PGDL competition winner scores at the bottom.
The highest score within each task is bolded.
In the CINIC-10 columns, `bn' stands batch-norm.
}
\label{tab:combo_norm_pca_gen_pred_results}
\begin{center}
\begin{widetabular}{\textwidth}{l | c c | c | c c | c | c | c | c}
\toprule 
\multicolumn{1}{c|}{}
&\multicolumn{2}{c|}{CIFAR-10}
&\multicolumn{1}{c|}{SVHN}
&\multicolumn{2}{c|}{CINIC-10}
&\multicolumn{1}{c|}{\makecell{Oxford \\ Flowers}}
&\multicolumn{1}{c|}{\makecell{Oxford \\Pets}}
&\multicolumn{1}{c|}{\makecell{Fashion \\ MNIST}}
&\makecell[t]{\textit{All} \\ \textit{Avg}}
\\\cline{2-9}
\multicolumn{1}{c|}{}
&\textit{VGG}
&\textit{NiN}
&\textit{NiN}
&\makecell{\textit{Conv}\\\textit{w/bn}}
&\makecell{\textit{Conv}\\}
&\textit{NiN}
&\textit{NiN}
&\textit{VGG}
\\ \midrule
NPCA Gi intra $\ell$0 \& Gi intra $\ell$1    & 0.31         & 27.49        & 17.92        & 16.13        & 12.12        & 15.65        & 10.08        & 4.83         & 13.07        \\ 
NPCA Gi intra $\ell$0 \& Gi inter $\ell$0    & 1.20         & 32.38        & 34.27        & 22.33        & 8.79         & 40.31        & 17.03        & 5.16         & 20.18        \\ 
NPCA Gi intra $\ell$0 \& Gi inter $\ell$1    & 2.68         & 29.00        & 19.63        & 18.41        & 9.64         & 23.20        & 10.71        & 11.93        & 15.65        \\ 
NPCA Gi intra $\ell$0 \& Pal intra $\ell$0   & 0.85         & 31.94        & \bf43.06     & 25.01        & 12.15        & \bf43.52     & 17.07        & 6.10         & 22.46        \\ 
NPCA Gi intra $\ell$0 \& Pal intra $\ell$1   & 0.34         & 28.03        & 17.68        & 16.54        & 12.63        & 16.07        & 10.99        & 6.44         & 13.59        \\ 
NPCA Gi intra $\ell$0 \& Pal inter $\ell$0   & 1.24         & 33.23        & 38.39        & 23.53        & 8.74         & 40.56        & 17.70        & 5.35         & 21.09        \\ 
NPCA Gi intra $\ell$0 \& Mix intra $\ell$0   & 0.21         & 33.05        & 41.86        & \bf30.29     & \bf18.87     & 42.47        & 16.08        & 7.78         & \bf23.83     \\ 
NPCA Gi intra $\ell$1 \& Gi inter $\ell$0    & 6.07         & 26.82        & 16.44        & 15.95        & 9.90         & 14.93        & 10.05        & 6.12         & 13.29        \\ 
NPCA Gi intra $\ell$1 \& Mix intra $\ell$0   & 0.07         & 29.17        & 16.78        & 22.69        & 18.45        & 15.39        & 10.40        & 8.22         & 15.15        \\ 
NPCA Gi inter $\ell$0 \& Gi inter $\ell$1    & \bf7.62      & 28.69        & 18.94        & 17.05        & 8.23         & 21.20        & 10.47        & 8.61         & 15.10        \\ 
NPCA Gi inter $\ell$0 \& Mix intra $\ell$0   & 1.79         & 35.07        & 39.08        & 30.05        & 15.40        & 40.30        & 16.87        & 8.03         & 23.33        \\ 
NPCA Gi inter $\ell$1 \& Mix intra $\ell$0   & 1.78         & 29.93        & 18.34        & 26.84        & 15.48        & 23.82        & 10.24        & \bf16.30     & 17.84        \\ 
NPCA Pal intra $\ell$0 \& Pal inter $\ell$0  & 1.38         & \bf35.39     & 41.46        & 25.20        & 9.51         & 40.44        & \bf18.18     & 7.28         & 22.36        \\ 
NPCA Pal intra $\ell$0 \& Pal inter $\ell$1  & 2.58         & 31.28        & 19.19        & 21.37        & 10.60        & 24.27        & 9.23         & 14.12        & 16.58        \\ 
\hline
\hline \hline
\textit{DBI*Mixup}                       & \it0.00      & \it25.86     & \it32.05     & \it31.79     & \it15.92     & \it43.99     & \it12.59     & \it9.24      & \it21.43     \\ 

\bottomrule
\end{widetabular}
\end{center}
\end{table}

\begin{table}[ht!]
\small
\caption{Comparison of Conditional Mutual Information scores for various complexity measures across tasks.
We present combinations of our measures using the simple average of two scores, and include the PGDL competition winner scores at the bottom.  Note in this case, since Gi and Pal scores are oppositely correlated, to take the simple average we use the negative of the Pal score added to the Gi score.
The highest score within each task is bolded.
In the CINIC-10 columns, `bn' stands batch-norm.
}
\label{tab:combo_avg_gen_pred_results}
\begin{center}
\begin{widetabular}{\textwidth}{l | c c | c | c c | c | c | c | c}
\toprule 
\multicolumn{1}{c|}{}
&\multicolumn{2}{c|}{CIFAR-10}
&\multicolumn{1}{c|}{SVHN}
&\multicolumn{2}{c|}{CINIC-10}
&\multicolumn{1}{c|}{\makecell{Oxford \\ Flowers}}
&\multicolumn{1}{c|}{\makecell{Oxford \\Pets}}
&\multicolumn{1}{c|}{\makecell{Fashion \\ MNIST}}
&\makecell[t]{\textit{All} \\ \textit{Avg}}
\\\cline{2-9}
\multicolumn{1}{c|}{}
&\textit{VGG}
&\textit{NiN}
&\textit{NiN}
&\makecell{\textit{Conv}\\\textit{w/bn}}
&\makecell{\textit{Conv}\\}
&\textit{NiN}
&\textit{NiN}
&\textit{VGG}
\\ \midrule
AVG Gi intra $\ell$0 \& Gi intra $\ell$1     & 0.06         & 26.25        & 18.26        & 16.16        & 12.09        & 15.77        & 7.80         & 5.10         & 12.69        \\ 
AVG Gi intra $\ell$0 \& Gi inter $\ell$0     & 1.42         & 32.94        & 34.64        & 22.25        & 8.87         & 38.66        & 17.05        & 4.49         & 20.04        \\ 
AVG Gi intra $\ell$0 \& Gi inter $\ell$1     & 4.29         & 27.89        & 20.38        & 18.11        & 9.64         & 17.69        & 8.73         & \bf15.88     & 15.33        \\ 
AVG Gi intra $\ell$0 \& Pal intra $\ell$0    & 0.89         & 35.65        & \bf43.53     & 26.67        & 12.86        & \bf43.43     & 18.00        & 6.94         & \bf23.50     \\ 
AVG Gi intra $\ell$0 \& Pal intra $\ell$1    & 0.24         & 15.71        & 12.59        & 10.03        & 13.09        & 7.15         & 5.22         & 8.35         & 9.05         \\ 
AVG Gi intra $\ell$0 \& Pal inter $\ell$0    & 3.28         & \bf37.21     & 32.59        & 23.89        & 7.71         & 34.35        & \bf19.94     & 5.36         & 20.54        \\ 
AVG Gi intra $\ell$0 \& Mix intra $\ell$0    & 0.13         & 16.33        & 20.56        & \bf30.17     & \bf18.15     & 29.83        & 7.87         & 13.22        & 17.03        \\ 
AVG Gi intra $\ell$1 \& Gi inter $\ell$0     & 4.43         & 26.76        & 16.42        & 16.14        & 9.86         & 20.90        & 8.86         & 4.70         & 13.51        \\ 
AVG Gi intra $\ell$1 \& Mix intra $\ell$0    & 0.40         & 9.15         & 22.71        & 23.89        & 9.28         & 18.83        & 1.83         & 10.92        & 12.13        \\ 
AVG Gi inter $\ell$0 \& Gi inter $\ell$1     & 7.63         & 27.03        & 19.48        & 16.99        & 8.25         & 20.86        & 8.84         & 7.37         & 14.56        \\ 
AVG Gi inter $\ell$0 \& Mix intra $\ell$0    & 2.99         & 7.67         & 19.00        & 24.59        & 18.04        & 15.05        & 5.01         & 0.98         & 11.66        \\ 
AVG Gi inter $\ell$1 \& Mix intra $\ell$0    & 0.91         & 3.66         & 26.51        & 19.92        & 10.61        & 7.48         & 0.78         & 6.53         & 9.55         \\ 
AVG Pal intra $\ell$0 \& Pal inter $\ell$0   & \bf24.84     & 29.70        & 14.04        & 1.64         & 3.45         & 14.84        & 2.13         & 4.89         & 11.94        \\ 
AVG Pal intra $\ell$0 \& Pal inter $\ell$1   & 17.74        & 7.09         & 7.04         & 0.69         & 1.08         & 0.47         & 2.03         & 14.88        & 6.38         \\ 
\hline
\hline \hline
\textit{DBI*Mixup$^1$}                       & \it0.00      & \it25.86     & \it32.05     & \it31.79     & \it15.92     & \it43.99     & \it12.59     & \it9.24      & \it21.43     \\ 

\bottomrule
\end{widetabular}
\end{center}
\end{table}

\begin{table}[ht!]
\small
\caption{Comparison of Conditional Mutual Information scores for various complexity measures across tasks.
We present combinations of our measures using the simple product of two scores, and include the PGDL competition winner scores at the bottom.  Note in this case, since Gi and Pal scores are oppositely correlated, we use the negative of the Pal score added to the Gi score.
The highest score within each task is bolded.
In the CINIC-10 columns, `bn' stands batch-norm.
}
\label{tab:combo_prod_gen_pred_results}
\begin{center}
\begin{widetabular}{\textwidth}{l | c c | c | c c | c | c | c | c}
\toprule 
\multicolumn{1}{c|}{}
&\multicolumn{2}{c|}{CIFAR-10}
&\multicolumn{1}{c|}{SVHN}
&\multicolumn{2}{c|}{CINIC-10}
&\multicolumn{1}{c|}{\makecell{Oxford \\ Flowers}}
&\multicolumn{1}{c|}{\makecell{Oxford \\Pets}}
&\multicolumn{1}{c|}{\makecell{Fashion \\ MNIST}}
&\makecell[t]{\textit{All} \\ \textit{Avg}}
\\\cline{2-9}
\multicolumn{1}{c|}{}
&\textit{VGG}
&\textit{NiN}
&\textit{NiN}
&\makecell{\textit{Conv}\\\textit{w/bn}}
&\makecell{\textit{Conv}\\}
&\textit{NiN}
&\textit{NiN}
&\textit{VGG}
\\ \midrule
PROD Gi intra $\ell$0 \& Gi intra $\ell$1    & 0.05         & 26.76        & 17.92        & 16.36        & \bf12.05     & 14.90        & 7.70         & 4.92         & 12.58        \\ 
PROD Gi intra $\ell$0 \& Gi inter $\ell$0    & 1.56         & 32.19        & 38.40        & 22.42        & 9.33         & \bf41.10     & 16.20        & 4.67         & 20.73        \\ 
PROD Gi intra $\ell$0 \& Gi inter $\ell$1    & 1.77         & 29.05        & 27.02        & 19.08        & 9.76         & 26.62        & 10.33        & 10.72        & 16.79        \\ 
PROD Gi intra $\ell$0 \& Pal intra $\ell$0   & 0.56         & 21.02        & 30.36        & 20.09        & 9.35         & 36.95        & 12.28        & 3.56         & 16.77        \\ 
PROD Gi intra $\ell$0 \& Pal intra $\ell$1   & 0.80         & 21.78        & 30.58        & 20.26        & 9.21         & 36.19        & 10.82        & 3.54         & 16.65        \\ 
PROD Gi intra $\ell$0 \& Pal inter $\ell$0   & 0.34         & 20.86        & 30.32        & 20.34        & 10.87        & 34.82        & 12.05        & 3.75         & 16.67        \\ 
PROD Gi intra $\ell$0 \& Mix intra $\ell$0   & 0.93         & 29.90        & 40.23        & 17.46        & 6.67         & 40.51        & 12.76        & 4.84         & 19.16        \\ 
PROD Gi intra $\ell$1 \& Gi inter $\ell$0    & 0.53         & 25.40        & 12.95        & 14.80        & 9.88         & 10.21        & 6.24         & 4.76         & 10.60        \\ 
PROD Gi intra $\ell$1 \& Mix intra $\ell$0   & 0.26         & 15.38        & 10.15        & 4.28         & 8.39         & 2.81         & 1.93         & 5.48         & 6.09         \\ 
PROD Gi inter $\ell$0 \& Gi inter $\ell$1    & 7.68         & 26.80        & 18.60        & 16.93        & 8.15         & 19.07        & 8.41         & 10.58        & 14.53        \\ 
PROD Gi inter $\ell$0 \& Mix intra $\ell$0   & 4.69         & 30.90        & 13.94        & 10.78        & 0.74         & 19.91        & 6.44         & 4.19         & 11.45        \\ 
PROD Gi inter $\ell$1 \& Mix intra $\ell$0   & \bf10.28     & 16.77        & 5.06         & 4.06         & 3.52         & 2.11         & 1.62         & 15.93        & 7.42         \\ 
PROD Pal intra $\ell$0 \& Pal inter $\ell$0  & 1.71         & \bf35.77     & \bf41.58     & \bf25.14     & 9.50         & 38.92        & \bf18.41     & 5.61         & \bf22.08     \\ 
PROD Pal intra $\ell$0 \& Pal inter $\ell$1  & 4.64         & 30.47        & 19.47        & 21.02        & 10.65        & 17.39        & 9.28         & \bf16.64     & 16.20        \\ 
\hline
\hline \hline
\textit{DBI*Mixup$^1$}                       & \it0.00      & \it25.86     & \it32.05     & \it31.79     & \it15.92     & \it43.99     & \it12.59     & \it9.24      & \it21.43     \\ 

\bottomrule
\end{widetabular}
\end{center}
\end{table}

\begin{table}[ht!]
\small
\caption{Comparison of Conditional Mutual Information scores for various complexity measures across tasks.
We present combinations of our measures using the product of two scores plus the average of the same two scores, and include the PGDL competition winner scores at the bottom.  Note in this case, since Gi and Pal scores are oppositely correlated, we use the negative of the Pal score added to the Gi score.
The highest score within each task is bolded.
In the CINIC-10 columns, `bn' stands batch-norm.
}
\label{tab:combo_prod_avg_gen_pred_results}
\begin{center}
\begin{widetabular}{\textwidth}{l | c c | c | c c | c | c | c | c}
\toprule 
\multicolumn{1}{c|}{}
&\multicolumn{2}{c|}{CIFAR-10}
&\multicolumn{1}{c|}{SVHN}
&\multicolumn{2}{c|}{CINIC-10}
&\multicolumn{1}{c|}{\makecell{Oxford \\ Flowers}}
&\multicolumn{1}{c|}{\makecell{Oxford \\Pets}}
&\multicolumn{1}{c|}{\makecell{Fashion \\ MNIST}}
&\makecell[t]{\textit{All} \\ \textit{Avg}}
\\\cline{2-9}
\multicolumn{1}{c|}{}
&\textit{VGG}
&\textit{NiN}
&\textit{NiN}
&\makecell{\textit{Conv}\\\textit{w/bn}}
&\makecell{\textit{Conv}\\}
&\textit{NiN}
&\textit{NiN}
&\textit{VGG}
\\ \midrule
PROD+AVG Gi intra $\ell$0 \& Gi intra $\ell$1& 0.06         & 26.26        & 18.24        & 16.16        & 12.10        & 15.73        & 7.78         & 5.10         & 12.68        \\ 
PROD+AVG Gi intra $\ell$0 \& Gi inter $\ell$0& 1.40         & 32.85        & 34.91        & 22.29        & 8.94         & 38.94        & 17.03        & 4.46         & 20.10        \\ 
PROD+AVG Gi intra $\ell$0 \& Gi inter $\ell$1& 4.17         & 27.90        & 20.69        & 18.28        & 9.63         & 18.42        & 8.89         & 15.70        & 15.46        \\ 
PROD+AVG Gi intra $\ell$0 \& Pal intra $\ell$0& 0.79         & 17.24        & 32.09        & \bf33.10     & \bf16.09     & \bf40.89     & 22.60        & 11.47        & 21.78        \\ 
PROD+AVG Gi intra $\ell$0 \& Pal intra $\ell$1& 2.44         & 6.87         & 3.59         & 1.41         & 12.05        & 0.12         & 3.10         & 1.34         & 3.86         \\ 
PROD+AVG Gi intra $\ell$0 \& Pal inter $\ell$0& \bf11.85     & \bf36.14     & 15.51        & 24.13        & 4.06         & 27.34        & \bf26.74     & 5.42         & 18.90        \\ 
PROD+AVG Gi intra $\ell$0 \& Mix intra $\ell$0& 0.77         & 4.65         & 12.16        & 14.22        & 10.46        & 23.94        & 5.03         & 13.96        & 10.65        \\ 
PROD+AVG Gi intra $\ell$1 \& Gi inter $\ell$0& 4.31         & 26.68        & 16.15        & 16.05        & 9.79         & 19.56        & 8.65         & 4.72         & 13.24        \\ 
PROD+AVG Gi intra $\ell$1 \& Mix intra $\ell$0& 0.64         & 1.54         & 13.75        & 8.81         & 1.40         & 11.97        & 0.89         & 11.33        & 6.29         \\ 
PROD+AVG Gi inter $\ell$0 \& Gi inter $\ell$1& 7.63         & 27.01        & 19.44        & 16.98        & 8.24         & 20.66        & 8.74         & 7.39         & 14.51        \\ 
PROD+AVG Gi inter $\ell$0 \& Mix intra $\ell$0& 8.53         & 1.82         & 12.87        & 6.86         & 11.41        & 3.96         & 2.26         & 2.29         & 6.25         \\ 
PROD+AVG Gi inter $\ell$1 \& Mix intra $\ell$0& 1.97         & 2.65         & 18.78        & 5.12         & 2.37         & 3.59         & 1.97         & 11.29        & 5.97         \\ 
PROD+AVG Pal intra $\ell$0 \& Pal inter $\ell$0& 1.75         & 35.79        & \bf41.46     & 25.12        & 9.45         & 38.82        & 18.42        & 5.59         & \bf22.05     \\ 
PROD+AVG Pal intra $\ell$0 \& Pal inter $\ell$1& 4.73         & 30.42        & 19.34        & 20.91        & 10.65        & 17.15        & 9.13         & \bf16.62     & 16.12        \\ 
\hline
\hline \hline
\textit{DBI*Mixup$^1$}                       & \it0.00      & \it25.86     & \it32.05     & \it31.79     & \it15.92     & \it43.99     & \it12.59     & \it9.24      & \it21.43     \\ 

\bottomrule
\end{widetabular}
\end{center}
\end{table}

\begin{table}[ht!]
\small
\caption{Comparison of Conditional Mutual Information scores for various complexity measures across tasks.
We present combinations of our measures using the average of two scores converted to ranks across the task (i.e., rank all the scores for the task for the particular score, and add $1$ go get the transformed score), and include the PGDL competition winner scores at the bottom.  Note in this case, since Gi and Pal scores are oppositely correlated, we use the negative of the Pal score added to the Gi score.
The highest score within each task is bolded.
In the CINIC-10 columns, `bn' stands batch-norm.
}
\label{tab:combo_avg_rank_gen_pred_results}
\begin{center}
\begin{widetabular}{\textwidth}{l | c c | c | c c | c | c | c | c}
\toprule 
\multicolumn{1}{c|}{}
&\multicolumn{2}{c|}{CIFAR-10}
&\multicolumn{1}{c|}{SVHN}
&\multicolumn{2}{c|}{CINIC-10}
&\multicolumn{1}{c|}{\makecell{Oxford \\ Flowers}}
&\multicolumn{1}{c|}{\makecell{Oxford \\Pets}}
&\multicolumn{1}{c|}{\makecell{Fashion \\ MNIST}}
&\makecell[t]{\textit{All} \\ \textit{Avg}}
\\\cline{2-9}
\multicolumn{1}{c|}{}
&\textit{VGG}
&\textit{NiN}
&\textit{NiN}
&\makecell{\textit{Conv}\\\textit{w/bn}}
&\makecell{\textit{Conv}\\}
&\textit{NiN}
&\textit{NiN}
&\textit{VGG}
\\ \midrule
AVG RANK Gi intra $\ell$0 \& Gi intra $\ell$1& 0.03         & 25.80        & 17.61        & 16.54        & 12.71        & 15.45        & 10.31        & 5.16         & 12.95        \\ 
AVG RANK Gi intra $\ell$0 \& Gi inter $\ell$0& 0.98         & 32.60        & 34.26        & 22.09        & 9.14         & 39.86        & 17.36        & 4.39         & 20.09        \\ 
AVG RANK Gi intra $\ell$0 \& Gi inter $\ell$1& 1.67         & 28.73        & 18.77        & 18.60        & 10.31        & 22.13        & 11.63        & \bf11.45     & 15.41        \\ 
AVG RANK Gi intra $\ell$0 \& Pal intra $\ell$0& 0.85         & 32.73        & \bf43.41     & \bf24.86     & 12.19        & \bf43.38     & 17.12        & 6.21         & \bf22.60     \\ 
AVG RANK Gi intra $\ell$0 \& Pal intra $\ell$1& 0.05         & 27.39        & 19.73        & 17.12        & \bf13.07     & 15.89        & 11.39        & 6.54         & 13.90        \\ 
AVG RANK Gi intra $\ell$0 \& Pal inter $\ell$0& 1.02         & \bf33.73     & 38.36        & 23.52        & 9.25         & 40.74        & \bf18.16     & 4.70         & 21.19        \\ 
AVG RANK Gi intra $\ell$0 \& Mix intra $\ell$0& 4.15         & 0.74         & 1.46         & 0.18         & 1.45         & 0.34         & 1.03         & 2.49         & 1.48         \\ 
AVG RANK Gi intra $\ell$1 \& Gi inter $\ell$0& 1.38         & 26.51        & 15.25        & 15.69        & 10.07        & 14.98        & 11.03        & 4.54         & 12.43        \\ 
AVG RANK Gi intra $\ell$1 \& Mix intra $\ell$0& 0.89         & 1.36         & 1.90         & 0.99         & 0.63         & 2.96         & 3.04         & 1.69         & 1.68         \\ 
AVG RANK Gi inter $\ell$0 \& Gi inter $\ell$1& 6.36         & 28.48        & 17.11        & 17.22        & 8.44         & 18.98        & 12.71        & 8.11         & 14.68        \\ 
AVG RANK Gi inter $\ell$0 \& Mix intra $\ell$0& 7.45         & 0.82         & 1.38         & 0.22         & 2.61         & 0.44         & 1.74         & 3.51         & 2.27         \\ 
AVG RANK Gi inter $\ell$1 \& Mix intra $\ell$0& 4.68         & 1.76         & 3.21         & 0.45         & 2.04         & 1.33         & 3.02         & 0.43         & 2.12         \\ 
AVG RANK Pal intra $\ell$0 \& Pal inter $\ell$0& \bf8.17      & 0.70         & 1.42         & 0.32         & 3.08         & 0.69         & 1.34         & 2.73         & 2.31         \\ 
AVG RANK Pal intra $\ell$0 \& Pal inter $\ell$1& 7.90         & 2.53         & 2.04         & 0.44         & 0.80         & 1.26         & 1.93         & 1.49         & 2.30         \\ 
\hline
\hline \hline
\textit{DBI*Mixup$^1$}                       & \it0.00      & \it25.86     & \it32.05     & \it31.79     & \it15.92     & \it43.99     & \it12.59     & \it9.24      & \it21.43     \\

\bottomrule
\end{widetabular}
\end{center}
\end{table}

\subsection{Measuring invariance: Additional experimental setup information}\label{app:invariance_exp}
In this section, we provide additional details about the measuring invariance experiments.

For training the Resnet and VGG networks on CIFAR-10 and SVHN, we rely on the the PyTorch framework \cite{NEURIPS2019_9015} and the PyTorch-Lightning wrapper \cite{falcon2019pytorch}.
In order to mimic a real world training paradigm, we split the training sets for both CIFAR-10 and SVHN into 95\% training data and 5\% validation data.

Each dataset, model, perturbation experiment combination is performed with 1 CPU and 1 V100 GPU.

\subsection{Measuring invariance: Additional results}\label{app:invariance_res}
In this section, we include scatter plots of model generalization gap vs. our statistics and the baselines from the measuring invariance experiments.

\begin{figure}
    \centering
    \includegraphics[width=0.45\linewidth]{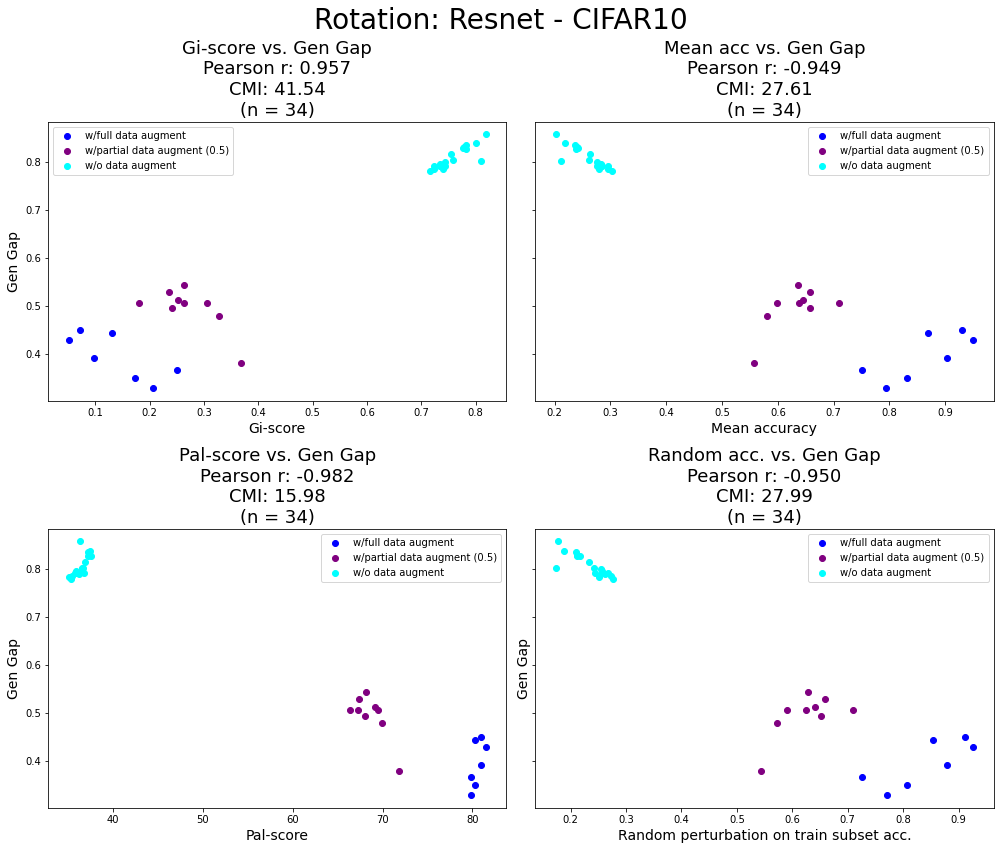}
    \vspace{3pt}
    \includegraphics[width=0.45\linewidth]{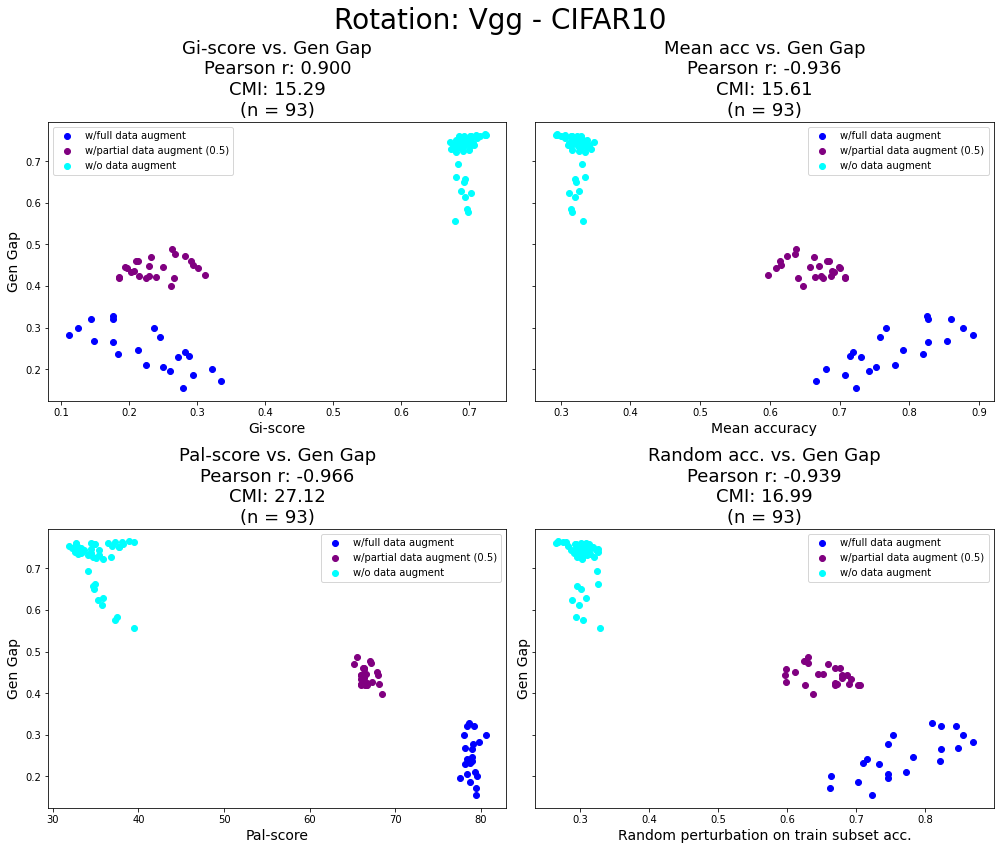}
    \vspace{3pt}
    \includegraphics[width=0.45\linewidth]{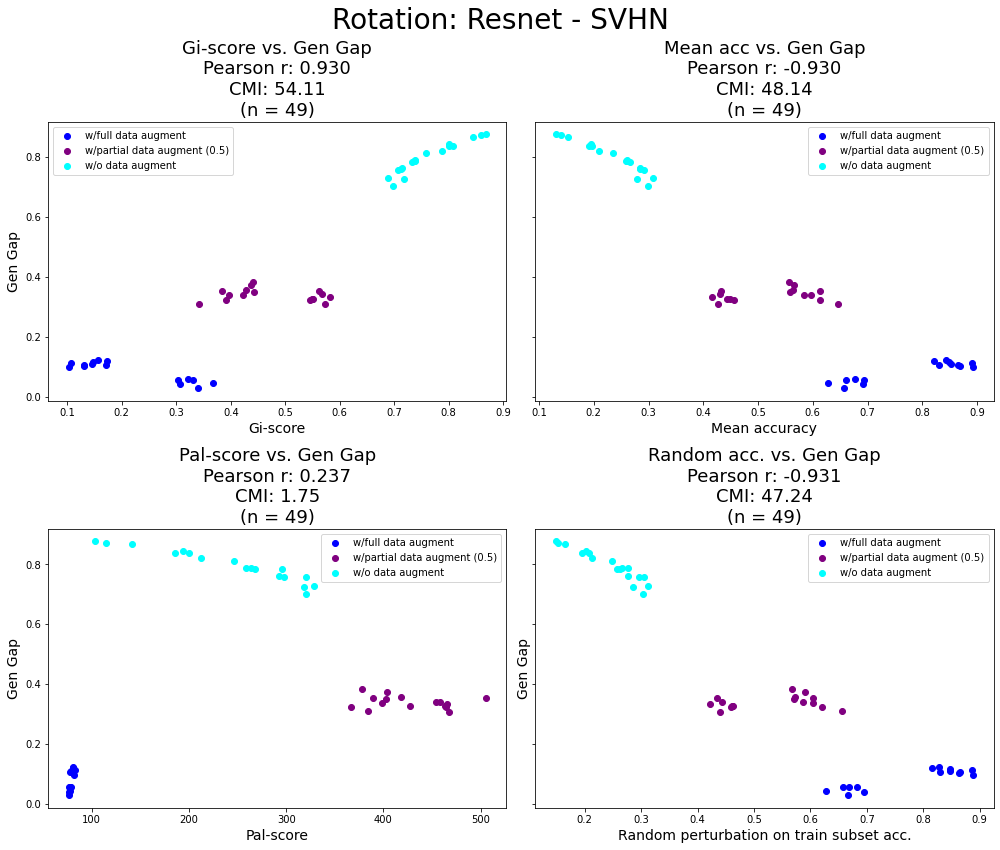}
    \vspace{3pt}
    \includegraphics[width=0.45\linewidth]{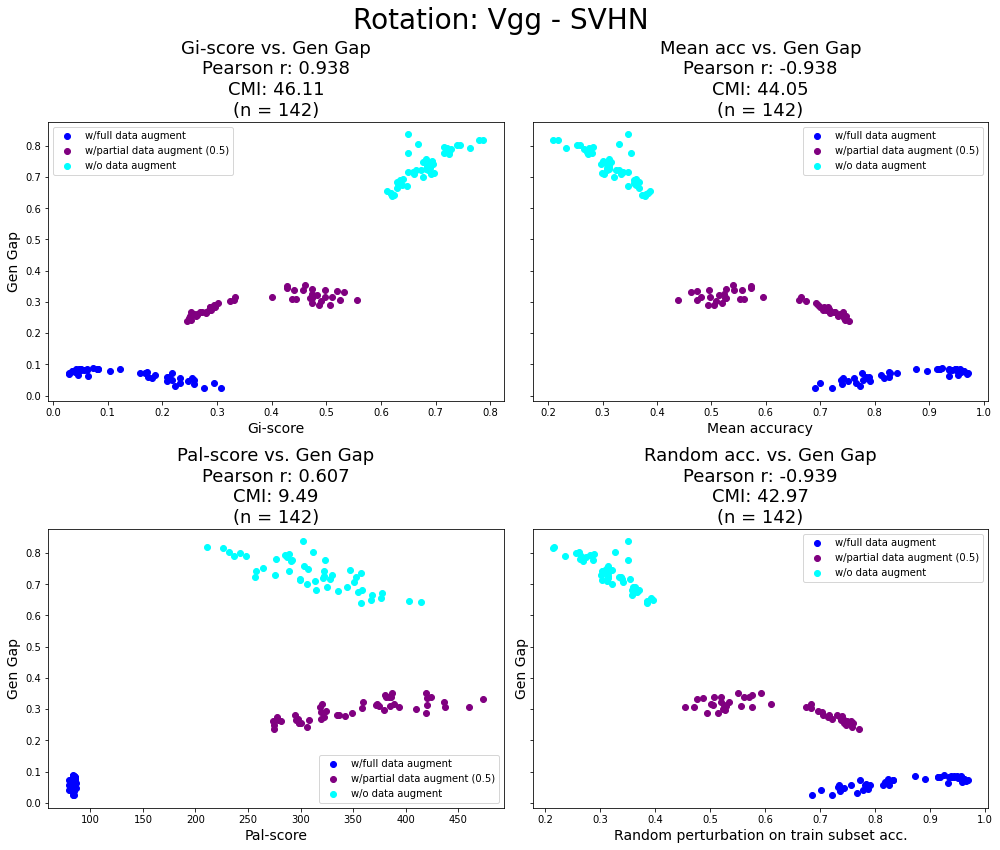}
    \caption{\textbf{Rotation:} Comparison of complexity measure and generalization gap for Resnet and VGG models trained on CIFAR-10 and SVHN to test how these measures predict generalization gap in the face of a rotation perturbation.}
    \label{fig:scatter_rotation}
\end{figure}

\begin{figure}
    \centering
    \includegraphics[width=0.45\linewidth]{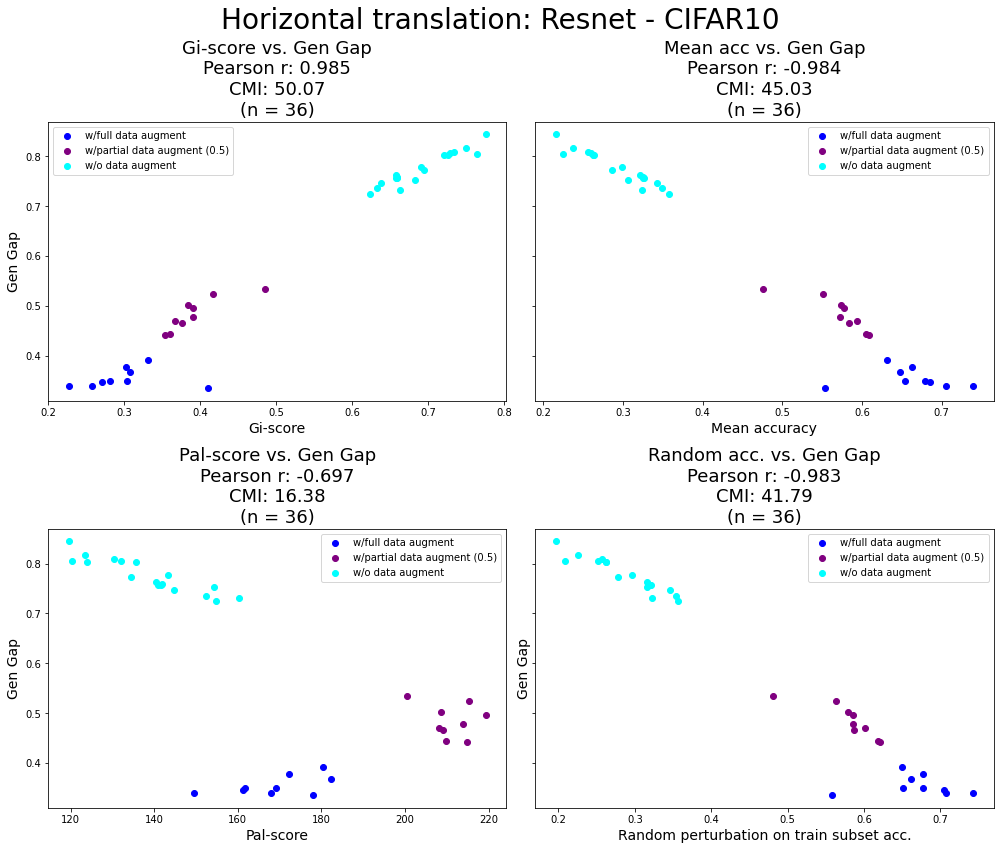}
    \vspace{3pt}
    \includegraphics[width=0.45\linewidth]{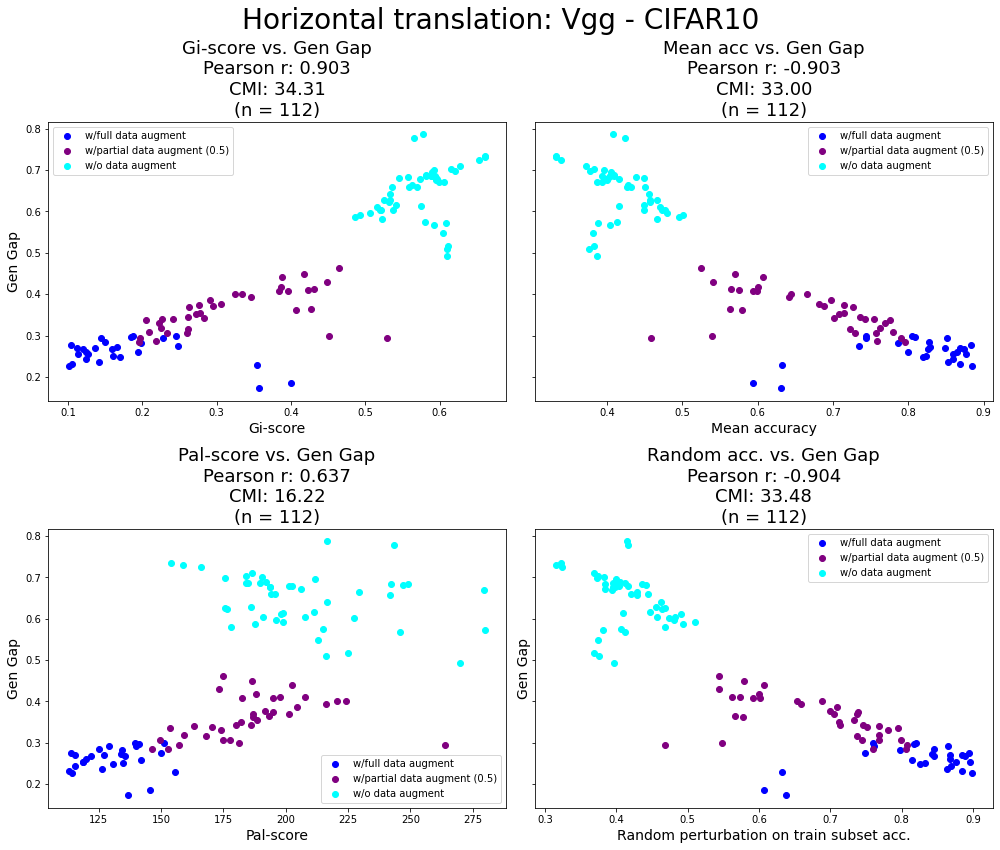}
    \vspace{3pt}
    \includegraphics[width=0.45\linewidth]{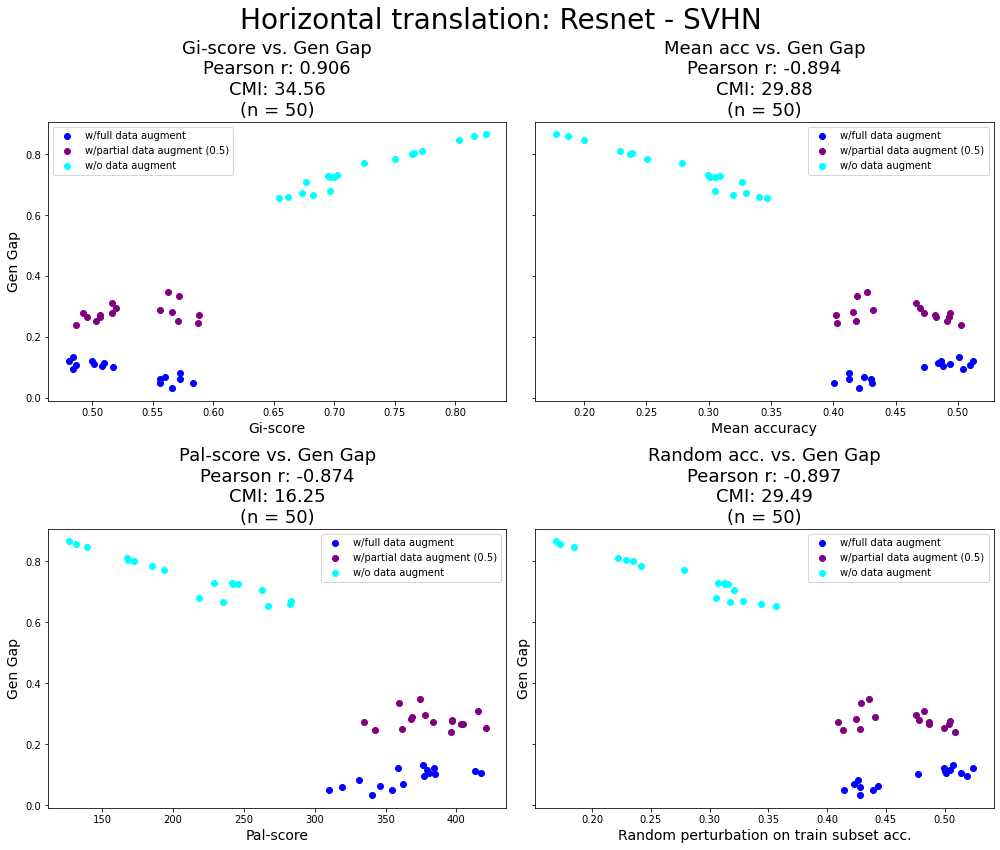}
    \vspace{3pt}
    \includegraphics[width=0.45\linewidth]{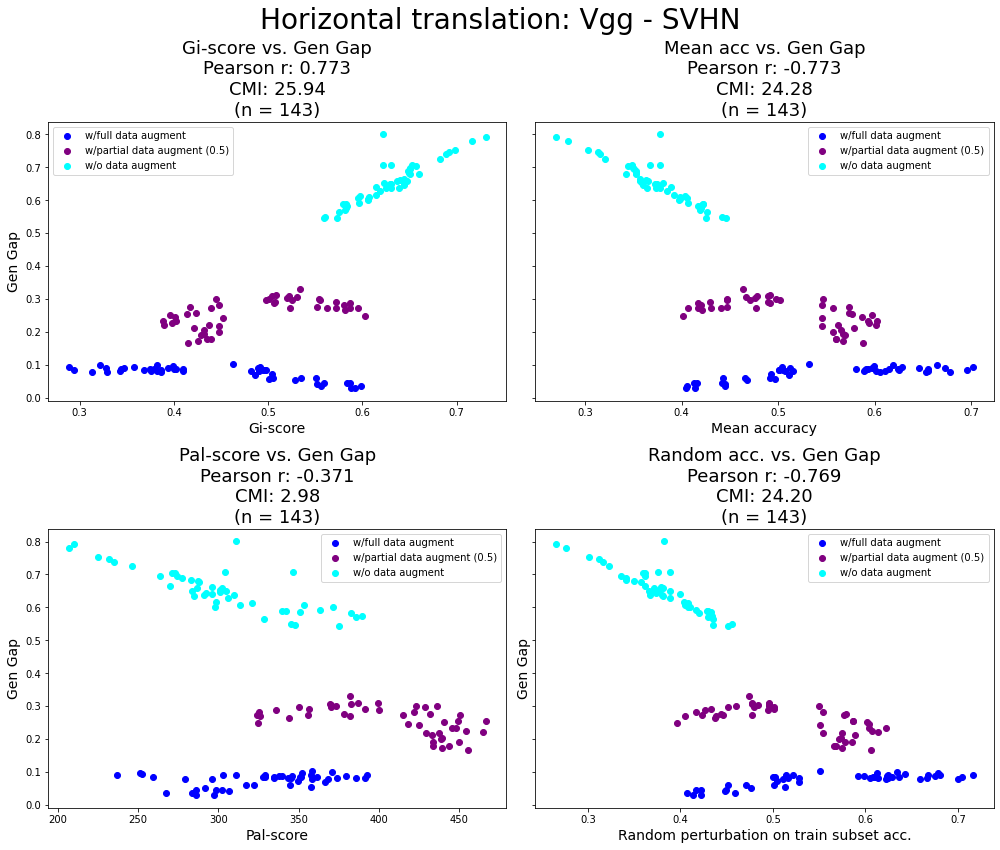}
    \caption{\textbf{Horizontal Translation:} Comparison of complexity measure and generalization gap for Resnet and VGG models trained on CIFAR-10 and SVHN to test how these measures predict generalization gap in the face of a horizontal translation perturbation.}
    \label{fig:scatter_horizontal_translation}
\end{figure}

\begin{figure}
    \centering
    \includegraphics[width=0.45\linewidth]{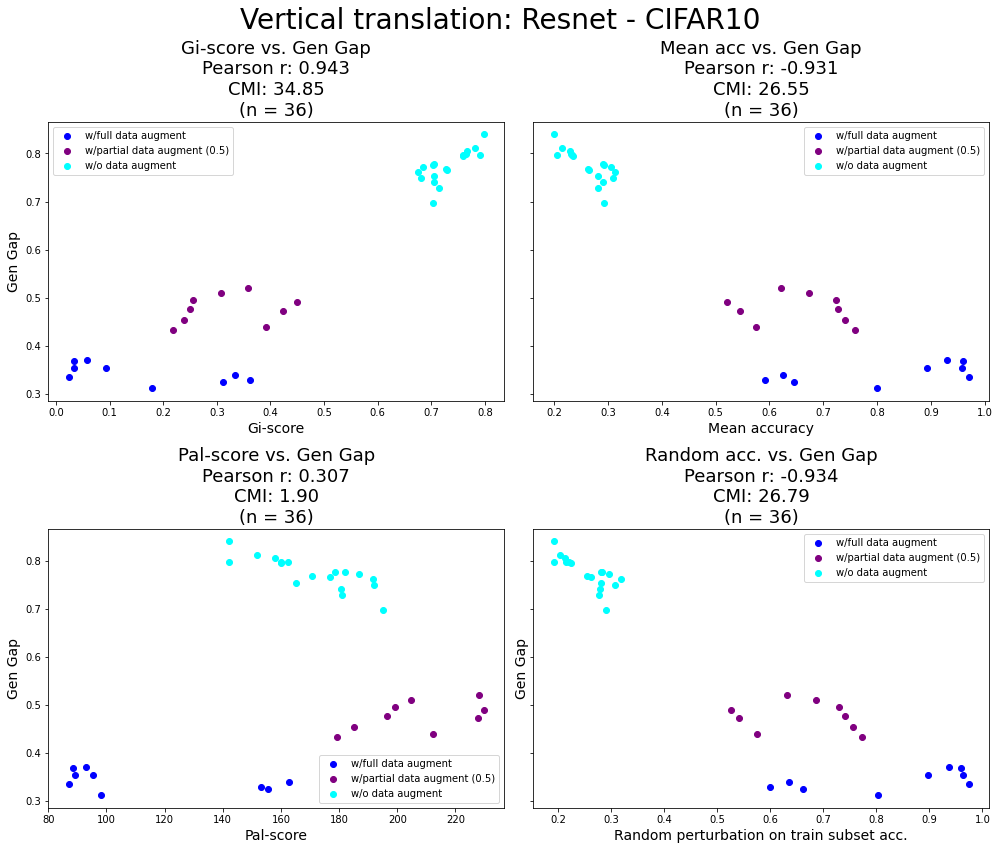}
    \vspace{3pt}
    \includegraphics[width=0.45\linewidth]{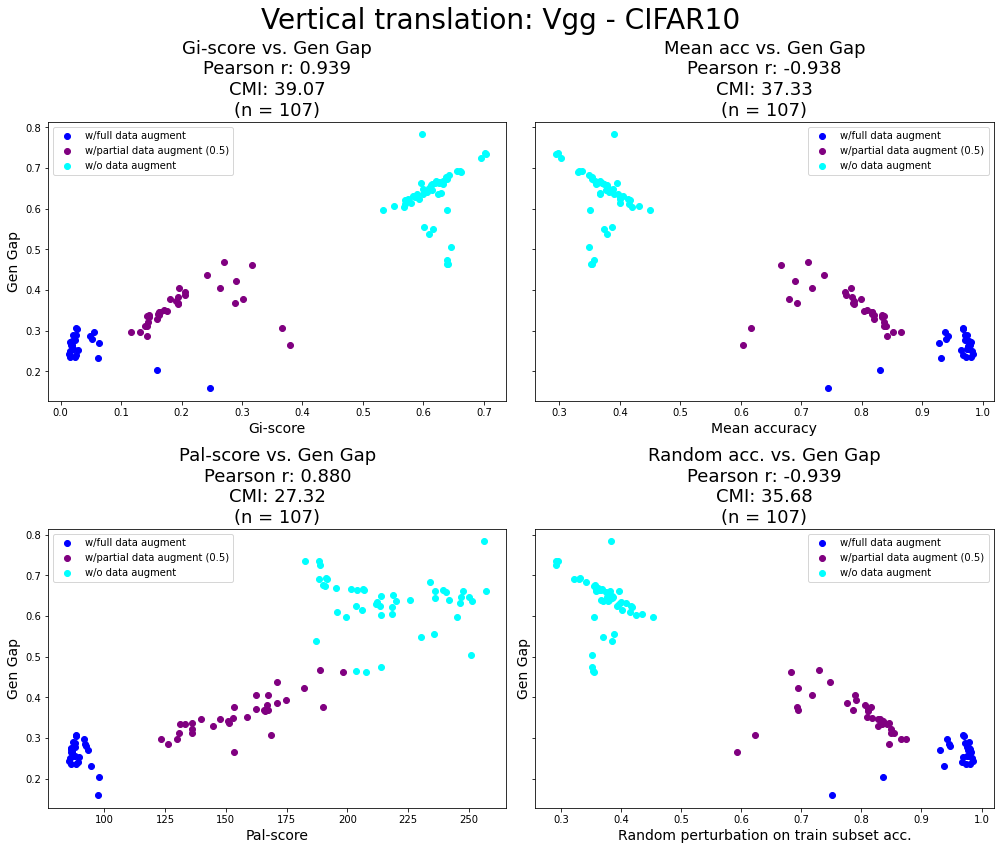}
    \vspace{3pt}
    \includegraphics[width=0.45\linewidth]{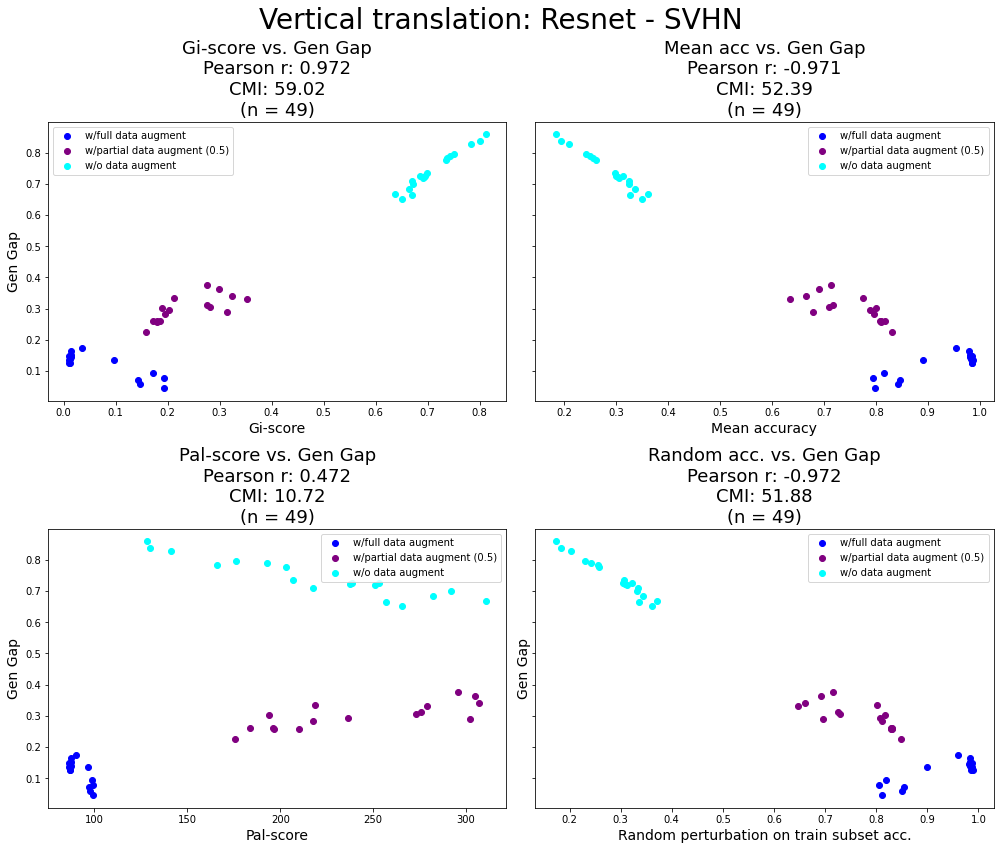}
    \vspace{3pt}
    \includegraphics[width=0.45\linewidth]{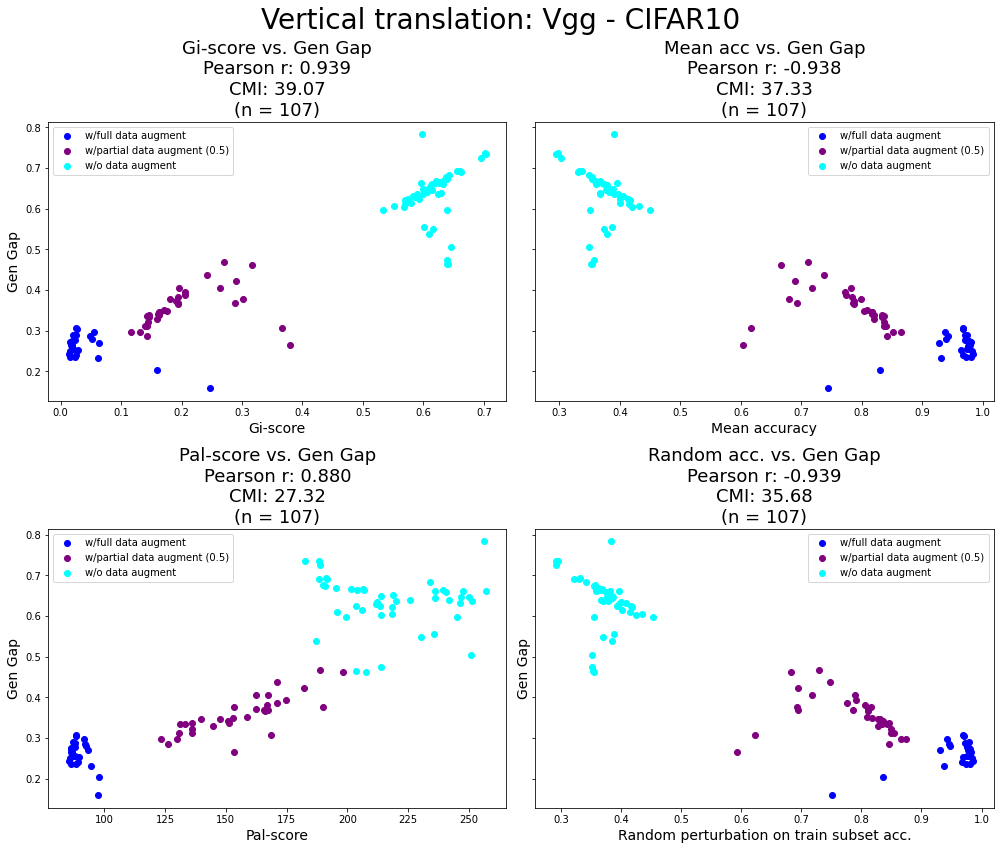}
    \caption{\textbf{Vertical Translation:} Comparison of complexity measure and generalization gap for Resnet and VGG models trained on CIFAR-10 and SVHN to test how these measures predict generalization gap in the face of a vertical translation perturbation.}
    \label{fig:scatter_vertical_translation}
\end{figure}

\begin{figure}
    \centering
    \includegraphics[width=0.45\linewidth]{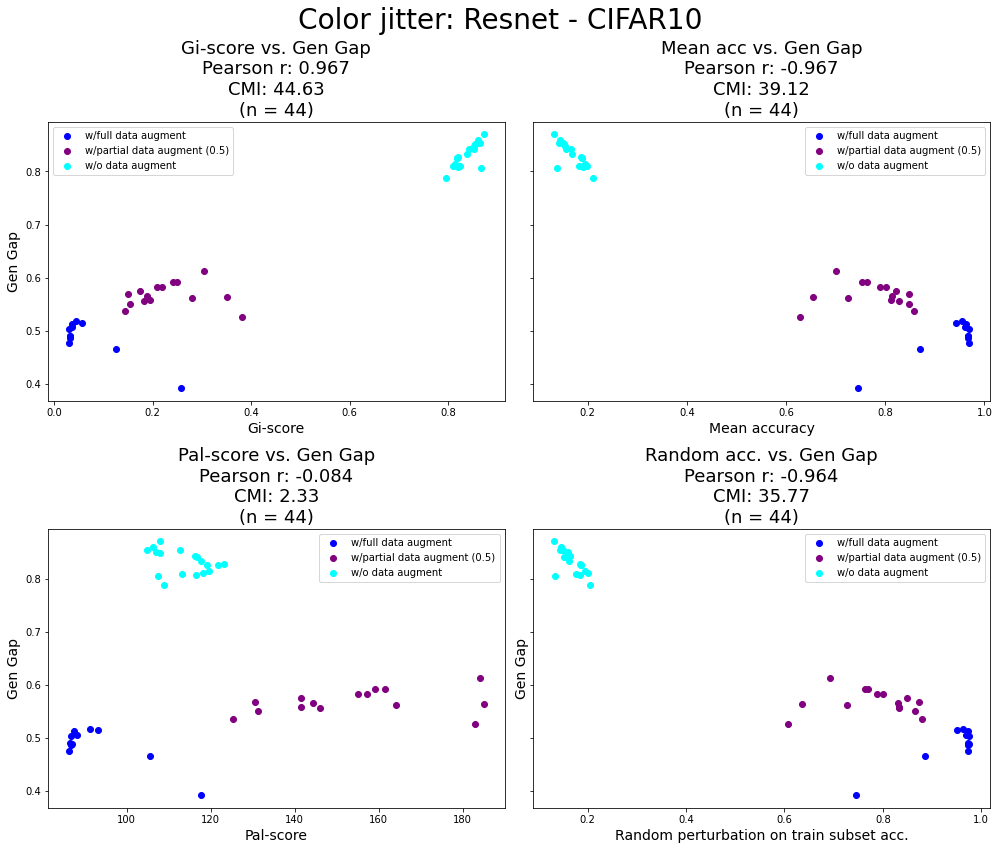}
    \vspace{3pt}
    \includegraphics[width=0.45\linewidth]{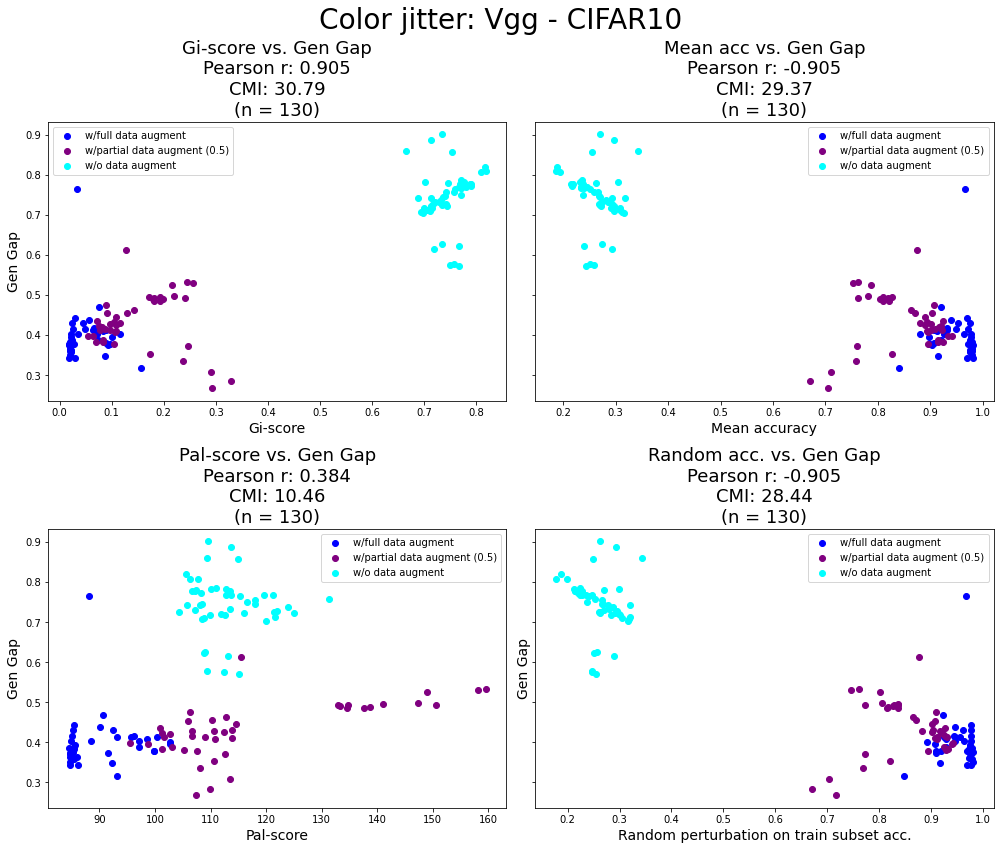}
    \vspace{3pt}
    \includegraphics[width=0.45\linewidth]{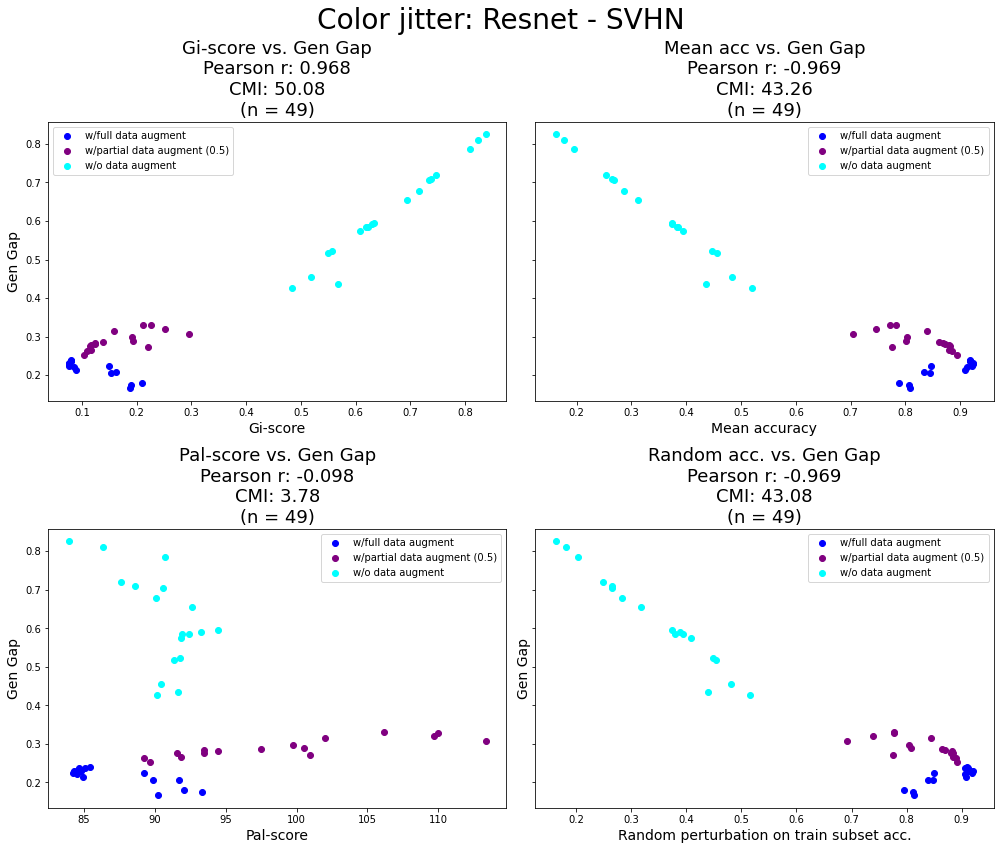}
    \vspace{3pt}
    \includegraphics[width=0.45\linewidth]{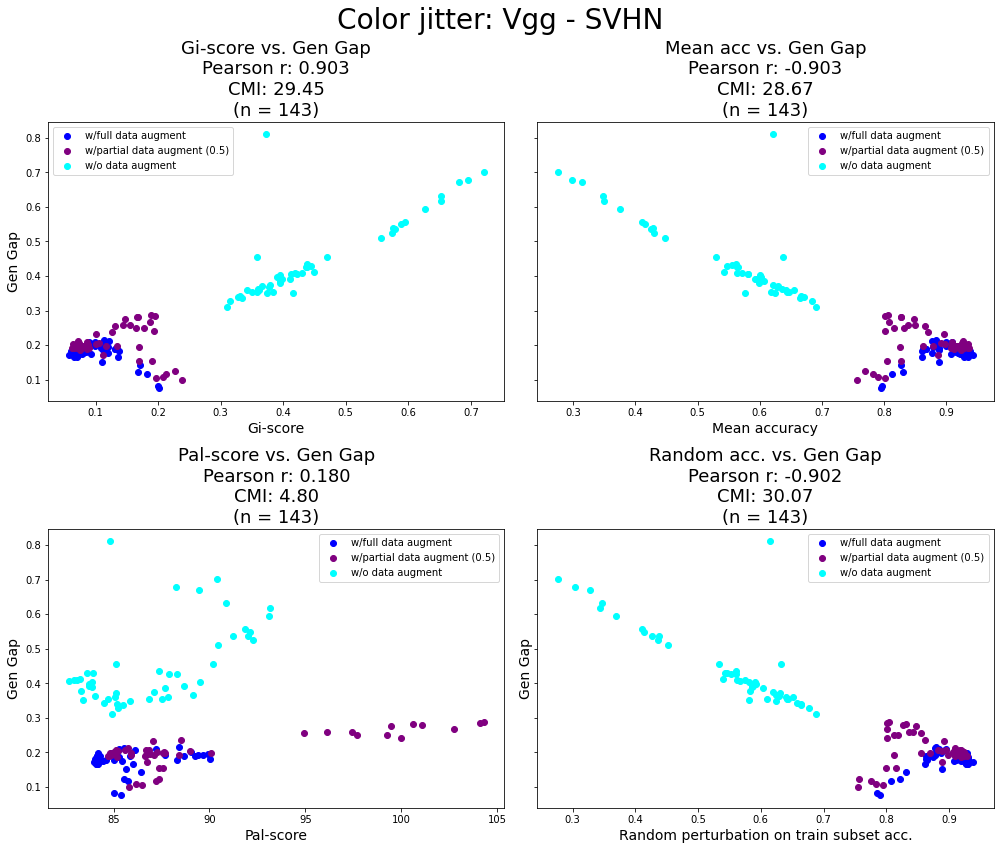}
    \caption{\textbf{Color-jittering:} Comparison of complexity measure and generalization gap for Resnet and VGG models trained on CIFAR-10 and SVHN to test how these measures predict generalization gap in the face of a color-jittering perturbation.}
    \label{fig:scatter_color_jitter}
\end{figure}

In each plot, we also include the number of models $n$, the CMI score, and the Pearson R correlation coefficient.
Results are displayed in Figures \ref{fig:scatter_rotation}, \ref{fig:scatter_horizontal_translation}, \ref{fig:scatter_vertical_translation}, and \ref{fig:scatter_color_jitter}.
Note that, interestingly, correlation and visual inspection do not always completely equate with CMI scores, as CMI specifically measures information provided by the complexity measure beyond what is known given the network hyperparameters / settings.
In particular, when the factors conditioned on correlate with generalization gap, the differentiating contribution of the complexity measure itself may not be as easily observable, and the CMI score is needed to fully elucidate the complexity measures' distinct predictive / informative capabilities.
We observe that our approach is best able to distinguish the different training condition groups and correlates most strongly with generalization gap, especially when augmented training is not applied so models are less likely to be invariant to the test perturbation (as opposed to when they are all trained to be invariant, which can potentially be achieved across hyperparameters if the perturbations are used in training).  

\end{document}